\UseRawInputEncoding
\pdfoutput=1 %采用pdflatex编译

\documentclass[11pt]{article}

\usepackage[english]{babel}
\usepackage{amssymb}
\usepackage{amsmath}
\usepackage{pdfpages}
\usepackage{caption}
\usepackage[letterpaper,top=2cm,bottom=2cm,left=3cm,right=3cm,marginparwidth=1.75cm]{geometry}

\usepackage{amsmath}
\usepackage{graphicx}
\usepackage[colorlinks=true, linkcolor=black, citecolor=blue]{hyperref}
\usepackage{color}
\usepackage{xcolor}
\usepackage{multirow}
\usepackage{booktabs}
\usepackage{xspace}
\usepackage[superscript]{cite}
\usepackage{float}

\usepackage{caption}  % For caption customization

\usepackage{enumitem}
\newcounter{suppfigure}
\makeatletter
\renewcommand{\maketitle}{\bgroup\setlength{\parindent}{0pt}
\begin{flushleft}
  \textbf{\@title}

  \@author
\end{flushleft}\egroup
}
\makeatother

\def\ourmethod{\texttt{Panacea}\xspace}
\def\ourbenchmark{\texttt{TrialPanorama}\xspace}
\def\ourdataalign{\texttt{TrialAlign}\xspace}
\def\ourdatainstr{\texttt{TrialInstruct}\xspace}

\begin{document}
\title{\ourmethod: A foundation model for clinical trial search, summarization, design, and recruitment}
\author{Jiacheng Lin$^{1}$, Hanwen Xu$^2$, Zifeng Wang$^1$, Sheng Wang$^{2\#}$, Jimeng Sun$^{1\#}$\\
$^1$ Department of Computer Science, University of Illinois Urbana-Champaign, Champaign, IL \\
$^2$ Paul G. Allen School of Computer Science and Engineering, University of Washington, Seattle, WA \\
  % \texttt{{zequnliu,mzhang\_cs}@pku.edu.cn}\\
  % \texttt{xuhw,swang@cs.washington.edu}
$^\#$Corresponding authors. Emails: swang@cs.washington.edu, jimeng@illinois.edu}

\maketitle
% \linenumbers

\begin{abstract}
Clinical trials are fundamental in developing new drugs, medical devices, and treatments. However, they are often time-consuming and have low success rates. Although there have been initial attempts to create large language models (LLMs) for clinical trial design and patient-trial matching, these models remain task-specific and not adaptable to diverse clinical trial tasks. To address this challenge, we propose a clinical trial foundation model named \ourmethod, designed to handle multiple tasks, including trial search, trial summarization, trial design, and patient-trial matching. We also assemble a large-scale dataset, named \ourdataalign, of 793,279 trial documents and 1,113,207 trial-related scientific papers, to infuse clinical knowledge into the model by pre-training. We further curate \ourdatainstr, which has 200,866 of instruction data for fine-tuning. 
These resources enable \ourmethod to be widely applicable for a range of clinical trial tasks based on user requirements.

We evaluated \ourmethod on a new benchmark, named \ourbenchmark, which covers eight clinical trial tasks. Our method performed the best on seven of the eight tasks compared to six cutting-edge generic or medicine-specific LLMs. Specifically, \ourmethod showed great potential to collaborate with human experts in crafting the design of eligibility criteria, study arms, and outcome measures, in multi-round conversations. In addition, Panacea achieved 14.42\% improvement in patient-trial matching, 41.78\% to 52.02\% improvement in trial search, and consistently ranked at the top for five aspects of trial summarization. Our approach demonstrates the effectiveness of \ourmethod in clinical trials and establishes a comprehensive resource, including training data, model, and benchmark, for developing clinical trial foundation models, paving the path for AI-based clinical trial development.
\end{abstract}

\newpage
\section*{Introduction}

Clinical trials are research studies conducted on humans to evaluate the safety and efficacy of new medical treatments, interventions, or devices before they are approved for widespread use. They form the foundation of modern medicine~\cite{ling2023clinical, heitmann2022covid, hammond2024phase, giamarellos2020activate, gilbert2022immune}. The challenges in clinical trials are three-fold. First, a clinical trial involves several interconnected design components, including trial descriptions, eligibility criteria, study arms, outcome metrics, and more, that need to be collectively designed to ensure optimal patient recruitment and outcome assessment. Second, clinical trial data are usually highly sensitive and private, hence often not amenable to pubic cloud-based tools (e.g., GPT-4~\cite{achiam2023gpt}) for processing and analysis. Third, clinical trial development requires multiple tasks, such as eligibility criteria design and patient recruitment, which require substantial domain expertise.

Machine learning models have shown promise in improving clinical trial development~\cite{DBLP:conf/emnlp/0010X023, gao2020compose, wang2022trial2vec, gligorijevic2019optimizing, zhang2020deepenroll, kim2021towards}. However, current models are often specialized for specific tasks, leading to challenges in managing the resulting models and utilizing training data effectively across interconnected clinical trial activities. Recently, foundation models have been highlighted as the generalist AI that can solve multiple tasks in many biomedical domains~\cite{DBLP:journals/corr/abs-2307-14334, moor2023foundation, lu2024visual, chen2024towards, cui2024scgpt, huang2023visual,xu2024whole}. For example, GPT-4 was used to assist clinical trial design and trial-patient matching~\cite{DBLP:journals/corr/abs-2307-15051, DBLP:conf/emnlp/0010X023, yuan2023large, DBLP:journals/corr/abs-2308-02180}. We thus hypothesize that a small but specialized clinical trial foundation model could be a Swiss Army Knife tool that simultaneously addresses multiple clinical trial tasks.

We present \ourmethod, a clinical trial foundation model that can address eight clinical trial tasks, including trial design, patient-trial matching, trial search, and trial summarization. The training of \ourmethod consists of an {\it alignment step} and an {\it instruction-tuning step}. During the alignment step, we train \ourmethod from a general-domain model using a large collection of trial documents and trial-related scientific papers. This step adapts \ourmethod to the vocabulary commonly used in clinical trials. To conduct the alignment, we create the \ourdataalign dataset from diverse resources, covering a comprehensive set of indications and medications for any clinical trial. The instruction-tuning step further enables \ourmethod to comprehend the user explanation of the task definition and the output requirement. By leveraging our curated \ourdatainstr dataset, \ourmethod can handle multiple clinical trial tasks without needing to re-train.

% In the instruction-tuning step, we teach \ourmethod how to solve each specific task. 
% Existing foundation models often fine-tune separately on different downstream applications, creating one model copy for each task. 
% However, because clinical trial tasks are highly interdependent, we propose to fine-tune our model using instruction data from all eight tasks simultaneously, creating a model that can address all tasks. 

We compared \ourmethod to six cutting-edge large language models on a new clinical trial benchmark \ourbenchmark. This benchmark covers eight tasks spanning trial design, patient-trial matching, trial search, and trial summarization. Our experiments showed that \ourmethod can facilitate experts through conversations, leading to superior design of eligibility criteria, study arms, and outcome measures. Especially on patient-trial matching, we found that \ our method achieved, on average, 14.42\% F1 improvement on two datasets. On trial search, \ourmethod obtained a 41.78\%  improvement in query generation and a 52.02\% improvement in query expansion. Finally, we propose evaluating trial summaries based on the alignment of their trial goals, conclusions, and keywords with reference summaries. We found that \ourmethod yield the best performance for the challenging multi-trial summarization tasks.

% Collectively, \ourmethod demonstrates a superior ability to solve various tasks in clinical trials, and together with \ourbenchmark, \ourdataalign, and \ourdatainstr, provides a complete framework for developing and evaluating clinical trial foundation models.
% On trial design, we found that \ourmethod outperformed existing models on designing criteria, study arms, and outcome measures. We further observed similar improvement on a De Novo trial design setting requiring minimal user inputs.

We have made all our training datasets (\ourdataalign and \ourdatainstr) and the evaluation benchmark (\ourbenchmark) available for future research and benchmarking of clinical trial foundation models. Additionally, we have open-sourced the code and model weights of \ourmethod. \ourmethod can run on a single-GPU machine, making it easy to use within an organization. Fine-tuning \ourmethod on 200 thousand documents only takes seven hours using a standard cluster with 4 A-100 GPUs. This advantage allows for further customization of \ourmethod on local proprietary data using limited computational resources.

\section*{Results}
\subsection*{Overview of \ourmethod}
Our goal is to develop  \ourmethod, a domain-specific foundation model for clinical trial tasks. Like previous works on developing domain-specific foundation models \cite{li2024llava, chaves2024training}, the biggest challenge for developing \ourmethod is to curate the high-quality fine-tuning data to align \ourmethod to clinical trial vocabulary and create the specific instruction data for clinical trial tasks.  
\ourmethod consists of two main steps: an alignment step, which adapts \ourmethod to the vocabulary used in clinical trials, and an instruction-tuning step, which instructs \ourmethod on each clinical trial task. We built two datasets \ourdataalign and \ourdatainstr for the alignment step and the instruction-tuning step, respectively.

%  and at least X conditions have more than 100,000 documents

\begin{figure}[!htbp]
    \centering
    \includegraphics[width=1.0\linewidth]{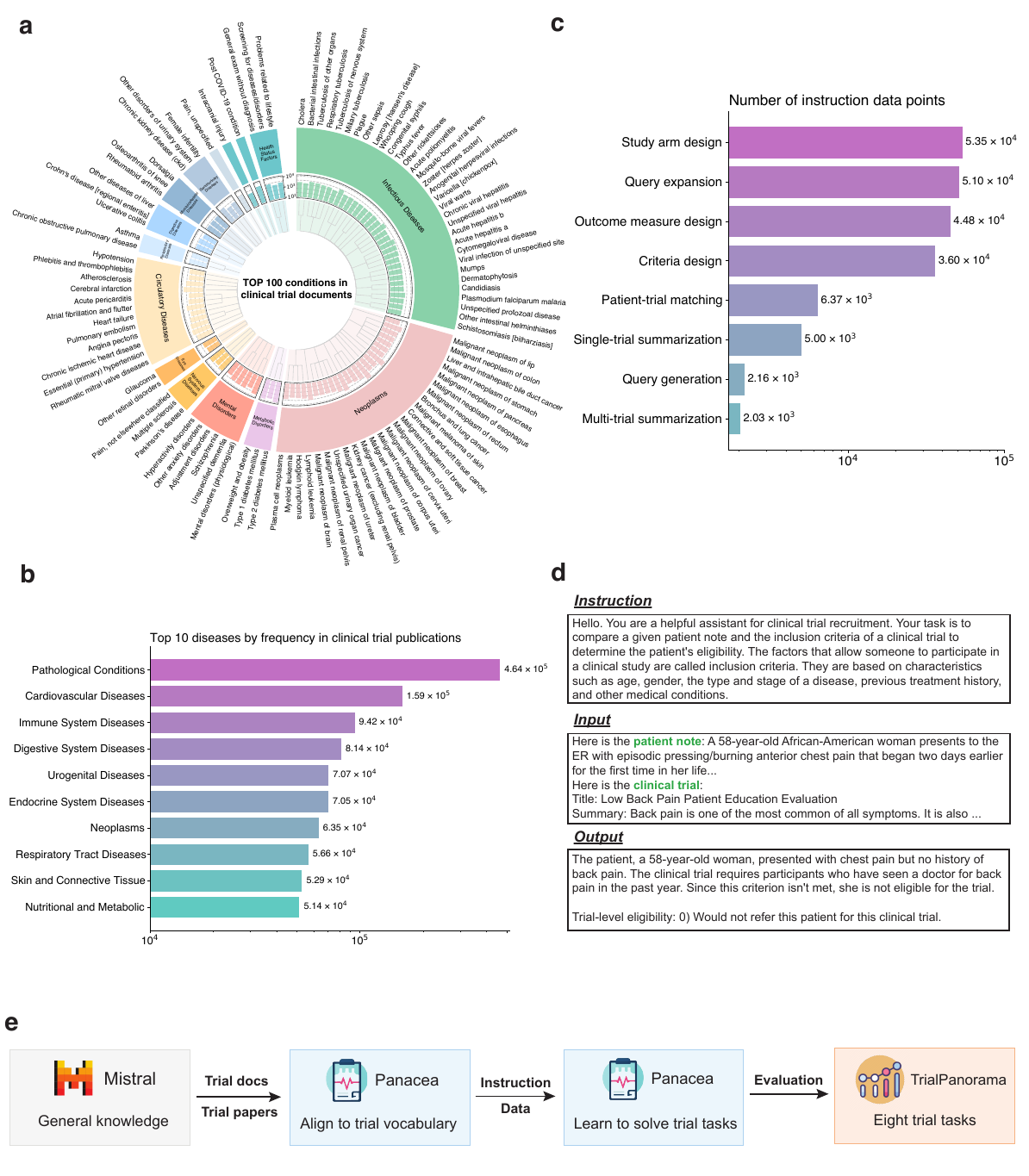} 
    \caption{\textbf{Overview of \ourmethod.}  \textbf{a,} Number of de-identified trial documents in each ICD-10 category. The top 100 conditions with the most number of trial documents are illustrated here. \textbf{b, } Bar plot showing the most frequent diseases in clinical trial publications according to the MeSH terms. \textbf{c, } Bar plot showing the number of instruction data points per clinical trial task in \ourdatainstr. \textbf{d, } An example of an instruction data point in \ourdatainstr. \textbf{e, } \ourmethod first uses \ourdataalign to fine-tune Mistral, then uses \ourdatainstr for instruction tuning. We create \ourbenchmark benchmark to evaluate \ourmethod and other LLMs on trial tasks. }
    \label{fig:1overview}
\end{figure}

\ourdataalign consists of 793,279 de-identified trial documents collected from 14 diverse sources and 1,113,207 scientific papers related to clinical trials (see \textbf{Methods}), representing a large-scale collection of clinical trial documents. By classifying these trial documents to terms in the International Classification of Diseases (ICD-10) ontology, we found that at least 100 conditions have 10,000 documents (\textbf{Fig. \ref{fig:1overview}a}), indicating the good coverage of our dataset. Likewise, by classifying trial-related scientific papers to Medical Subject Headings (MeSH) terms, we found that at least 119 terms have more than 10,000 papers and at least 1,921 terms have more than 1,000 papers  (\textbf{Fig. \ref{fig:1overview}b}). The scale and the coverage of \ourdataalign enable \ourmethod to be generalized to various conditions and treatments. 

\ourdatainstr contains instruction-tuning data from eight diverse tasks, including criteria design, study arm design, outcome measure design, patient-trial matching, query generation, query expansion, single-trial summarization, and multi-trial summarization, instructing \ourmethod on solving these tasks (\textbf{Fig. \ref{fig:1overview}c}). Each task contains at least 2,000 data points, where each data point contains an instruction, an input, and an output (\textbf{Fig. \ref{fig:1overview}d}). Since these eight tasks are related, we jointly fine-tuned the model using instruction data from these eight tasks, transforming \ourmethod into an all-in-one tool for clinical trial applications (\textbf{Fig. \ref{fig:1overview}e}). 

To evaluate \ourmethod, we built the first large-scale benchmark \ourbenchmark that covers eight specific tasks in clinical trials (\textbf{Table \ref{tbl:trialbench}}). Since these tasks contain both classification and generation tasks, \ourbenchmark allows us to evaluate \ourmethod in various machine learning settings. We made this benchmark fully open-source.

\begin{table}
\centering
\caption{We curate \ourbenchmark benchmark to evaluate our trial foundation \ourmethod on eight clinical trial tasks spanning trial design, patient-trial matching, trial search, and trial summarization. Here is the summary of the clinical trial tasks, dataset sizes, and evaluation metrics.}
\resizebox{\linewidth}{!}{
\begin{tabular}{llp{0.25\columnwidth}p{0.6\columnwidth}ccc}
\toprule
\multirow{2}{*}{\textbf{Task type}}           & \multirow{2}{*}{\textbf{Task name}} & \multirow{2}{*}{\textbf{Metric}} & \multirow{2}{*}{\textbf{Description}} & \multicolumn{3}{c}{\textbf{Data size}} \\
\cline{5-7}
     &            &                &                              & \textbf{Train}     & \textbf{Dev}     & \textbf{Test}    \\
     \midrule
\multirow{2}{*}{Trial search}        & Query generation     & Jaccard index    & Generate searchable queries based on specific clinical trial requirements for database retrieval.  & 1,837 & 324 &  925  \\
& Query expansion   &  Jaccard index   &  Broaden search parameters to include related terms and conditions to enhance trial discovery. & 43,350 & 7,650 & 2,500 \\
\midrule
\multirow{2}{*}{Trial summarization} & Single-trial summarization &  ROUGE, LLM-based metric &    Summarize key details and results of individual clinical trials.  & 4,250 & 7,50 & 1,000  \\
     & Multi-trial summarization  & ROUGE, LLM-based metric &     Compile and compare outcomes across multiple clinical trials for comprehensive insights.  & 1,725 & 304 & 252 \\
\midrule
\multirow{3}{*}{Trial design}        & Criteria design   &  \multirow{3}{*}{\begin{tabular}[c]{@{}l@{}}BLEU\\ ROUGE\\ Clinical relevance\end{tabular}}       & Define eligibility criteria for patient selection in clinical trials. & 30,559 & 5,392 & 549 \\
& Study arm design       &    &       Develop different intervention groups to assess the effects of treatments.  & 45516 & 8032 & 549    \\
& Outcome measure design   &  &      Establish methods for measuring trial results and effectiveness of interventions. & 38,088 & 6,721 & 549\\
\midrule
Patient-trial matching               & Patient-trial matching     &  F1, BACC, KAPPA &     Match eligible patients with suitable clinical trials, 3-class classification problem & 24,146 & 4,261 & 11,341\\
\bottomrule
\end{tabular}
}
    \label{tbl:trialbench}
\end{table}

\subsection*{Accurate trial search through query generation and expansion}
\begin{figure}[!htbp]
    \centering
    \includegraphics[width=1.0\linewidth]{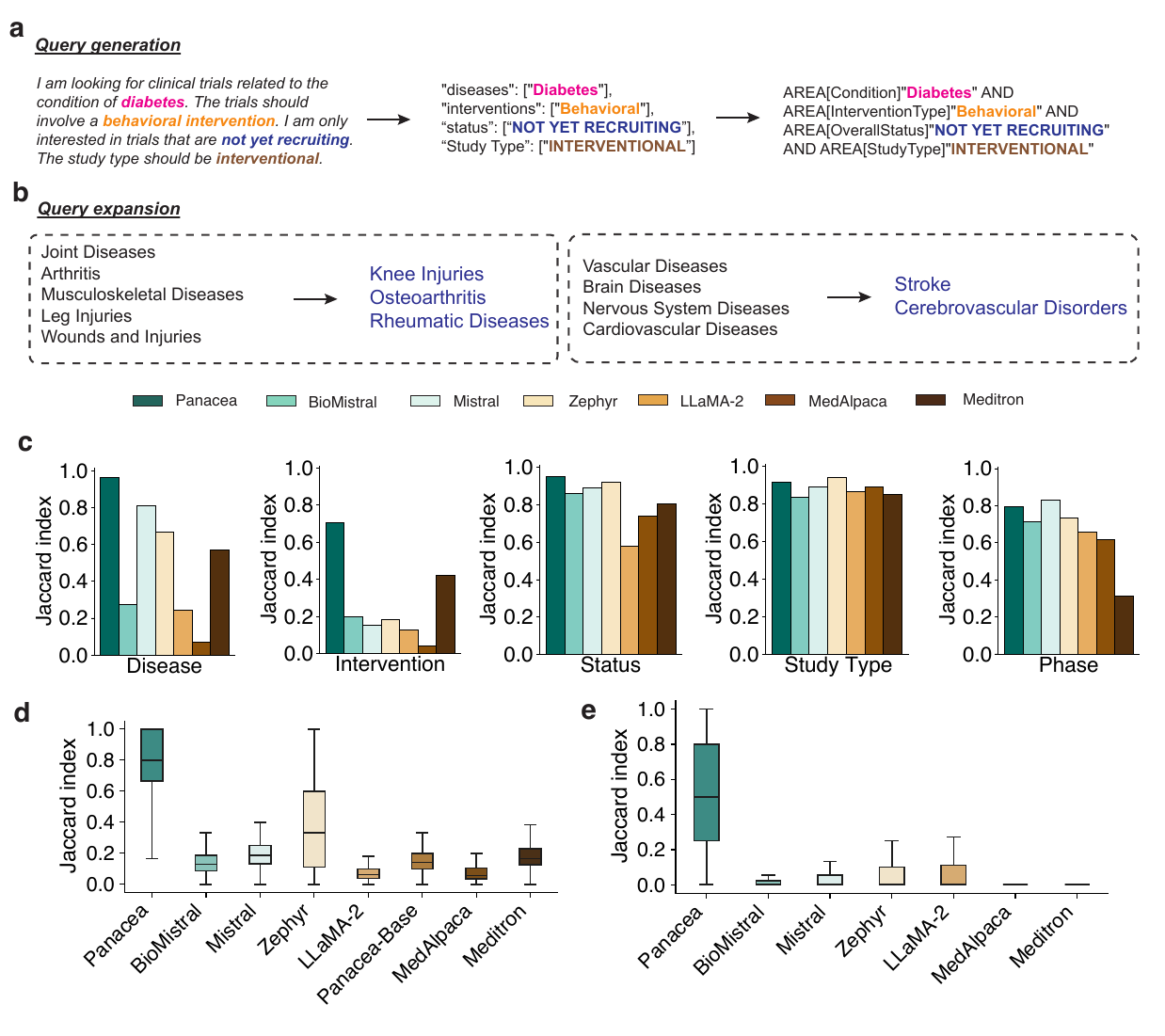} 
    \caption{\textbf{Evaluation on trial search.}  \textbf{a, } Query generation aims to convert free text user input into a structured query that contains five categories: disease, intervention, phase, status, and study type. \textbf{b, } Query expansion aims to expand a set of keywords. Candidate keywords are not provided. \textbf{c, } Comparison of query generation in five specific categories in terms of Jaccard index. \textbf{d, } Comparison of query generation in terms of Jaccard index.  \textbf{e, } Comparison of query expansion in terms of Jaccard index. }
    \label{fig:4search}
\end{figure}

Clinical trial search is an important task for clinical trial design and research.  Trial designers often need to study similar trials to ensure their design aligns with existing trials. 
The goal of the trial search is to find relevant trials based on user inputs, which serves as the foundation for designing and matching trials. The key to a successful trial search is to create comprehensive search terms. 
As a result, we evaluate {\it query generation}, which converts unstructured user input to a list of keywords (\textbf{Fig. \ref{fig:4search}a}), and {\it query expansion}, which further expands this keyword list to relevant terms (\textbf{Fig. \ref{fig:4search}b}). These two tasks assess the ability to derive high-quality queries based on user intent, which is crucial for a successful trial search.

We first evaluated query generation by formulating it as a text classification problem that classifies user inputs into specific diseases, interventions, phases, status, and study types. We found that \ourmethod substantially outperformed existing approaches regarding the Jaccard index (\textbf{Fig. \ref{fig:4search}d}). The improvement is larger on diseases and interventions, which are more challenging due to the large number of classes in these two categories (\textbf{Fig. \ref{fig:4search}c}), indicating that \ourmethod can accurately convert user inputs into the structured format that is compatible with downstream machine learning classifiers. 

Next, we evaluated query expansion by formulating it as a text generation problem. We did not provide the candidate keywords to the models since real-world keywords might have never been seen in the training trials. Similar to our observations in the query generation, \ourmethod achieved the best results on query expansion in terms of Jaccard index (\textbf{Fig. \ref{fig:4search}e}). We attribute the inferior performance of existing models on query expansion to the lack of fine-tuning on trial-related datasets. In contrast, \ourmethod is fine-tuned on \ourdataalign, adapting it to the vocabulary used in clinical trials. 
The promising results of \ourmethod on query expansion and generation demonstrate its ability to precisely understand user intent, providing an accurate tool for finding relevant clinical trials.

\subsection*{A novel metric to evaluate trial summarization}
\begin{figure}[!htbp]
    \centering
    \includegraphics[width=1.0\linewidth]{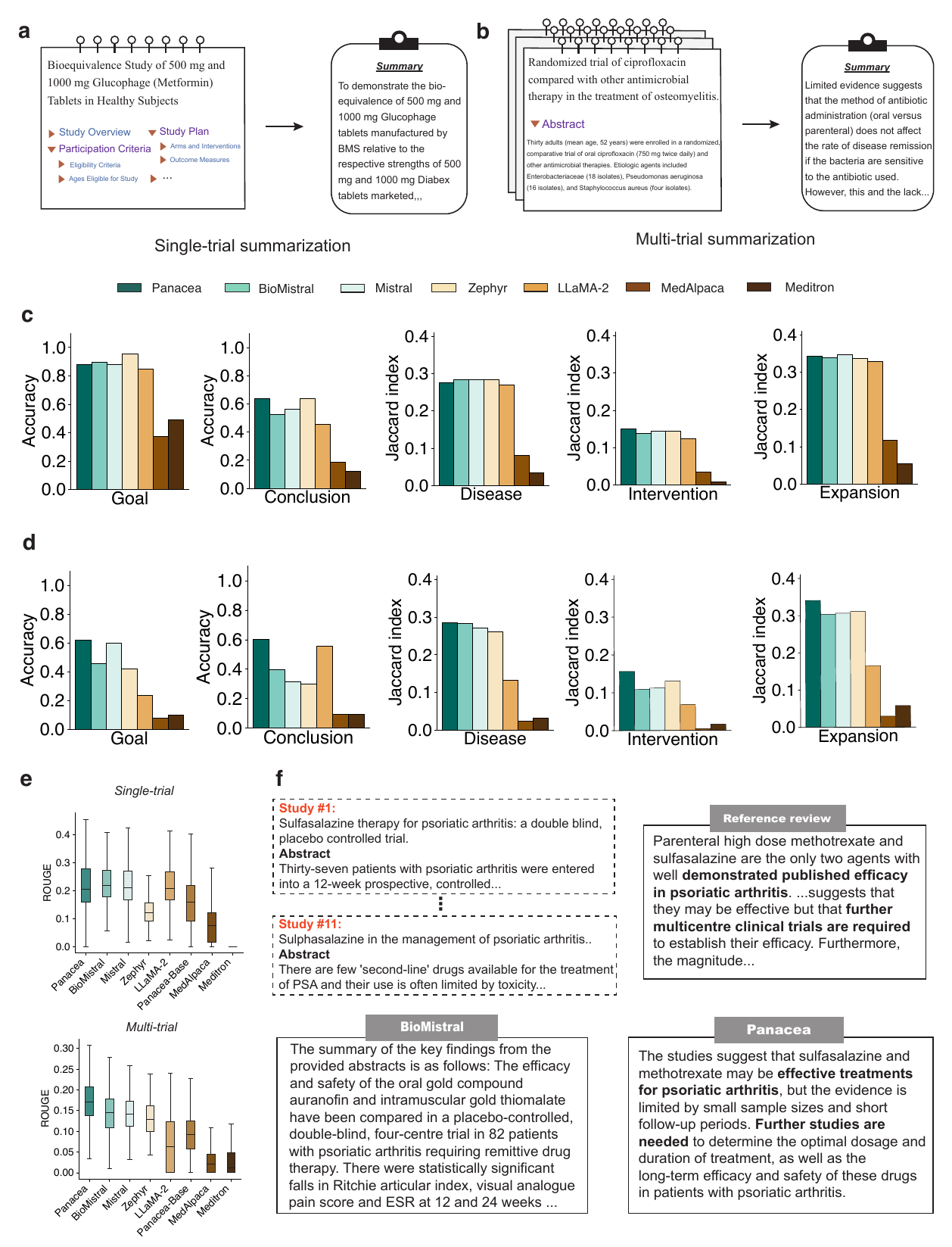} 
 \caption{\textbf{Evaluating \ourmethod on trial summarization.}  \textbf{a,b,} Trial summarization aims to provide a concise summary, including trial goal and conclusion, for a single trial (\textbf{a}) or multiple trials (\textbf{b}). \textbf{c,d, } Evaluation on single-trial summarization (\textbf{c}) and multiple-trial summarization (\textbf{d}) by using Claude-based metric and trial search-based metrics. \textbf{e, } Comparison on trial search in terms of ROUGE. \textbf{f, } A case study illustrating how \ourmethod successfully summarize multiple studies.}
    \label{fig:5summary}
\end{figure}

Once similar trials are identified, the next task is to understand those trials via summarization. 
We evaluated the performance of \ourmethod on trial summarization. We studied both single-trial summarization, which aims to provide a concise summary of a specific trial study (\textbf{Fig. \ref{fig:5summary}a}), and multi-trial summarization, which aims to summarize multiple trial studies that study similar conditions and interventions (\textbf{Fig. \ref{fig:5summary}b}).

Since it could be biased to evaluate summarization using lexical-based metrics, we propose a novel metric based on large language models (see \textbf{Methods}, \textbf{Supplementary Figures \ref{supfig:summary_eval_gpt} and \ref{supfig:summary_eval_gpt_multi}}). In particular, we provided the ground truth summarization and the model-generated summarization to Claude and asked if these summarizations studied the same problem and made the same conclusion. We found that \ourmethod and comparison approaches can correctly summarize the trial goal, while the summarization of the trial conclusion is less accurate (\textbf{Fig. \ref{fig:5summary}c-d}). Moreover, summarizing multiple trials is more challenging than summarizing a single trial based on the proposed metric. Nevertheless, our method still outperformed comparison approaches in summarizing multiple trials, suggesting its potential to assist researchers in extracting key information from many related trial studies. 

We further used query generation and query expansion to evaluate trial summarization by extracting diseases, and interventions, and expanding them (\textbf{Fig. \ref{fig:5summary}c-d}) from each trial. We examined whether the generated summarization can derive the same keywords as the ground truth summarization. We found that \ourmethod achieved the best performance on three of the six keyword categories while achieving comparable on the other categories. Moreover, we calculated the ROUGE score, which is used as the metric for trial summarization in previous works \cite{deyoung2021ms2, jiang2024trisum}, and observed improved performance by \ourmethod as well on multi-trial summarization (\textbf{Fig. \ref{fig:5summary}e}). Finally, we used a case study to show that \ourmethod can correctly summarize the goal and the conclusion for 11 trial studies, while comparison models failed to (\textbf{Fig. \ref{fig:5summary}f}).

\subsection*{Improved performance on clinical trial design}
\begin{figure}[!htbp]
    \centering
    \includegraphics[width=.95\linewidth]{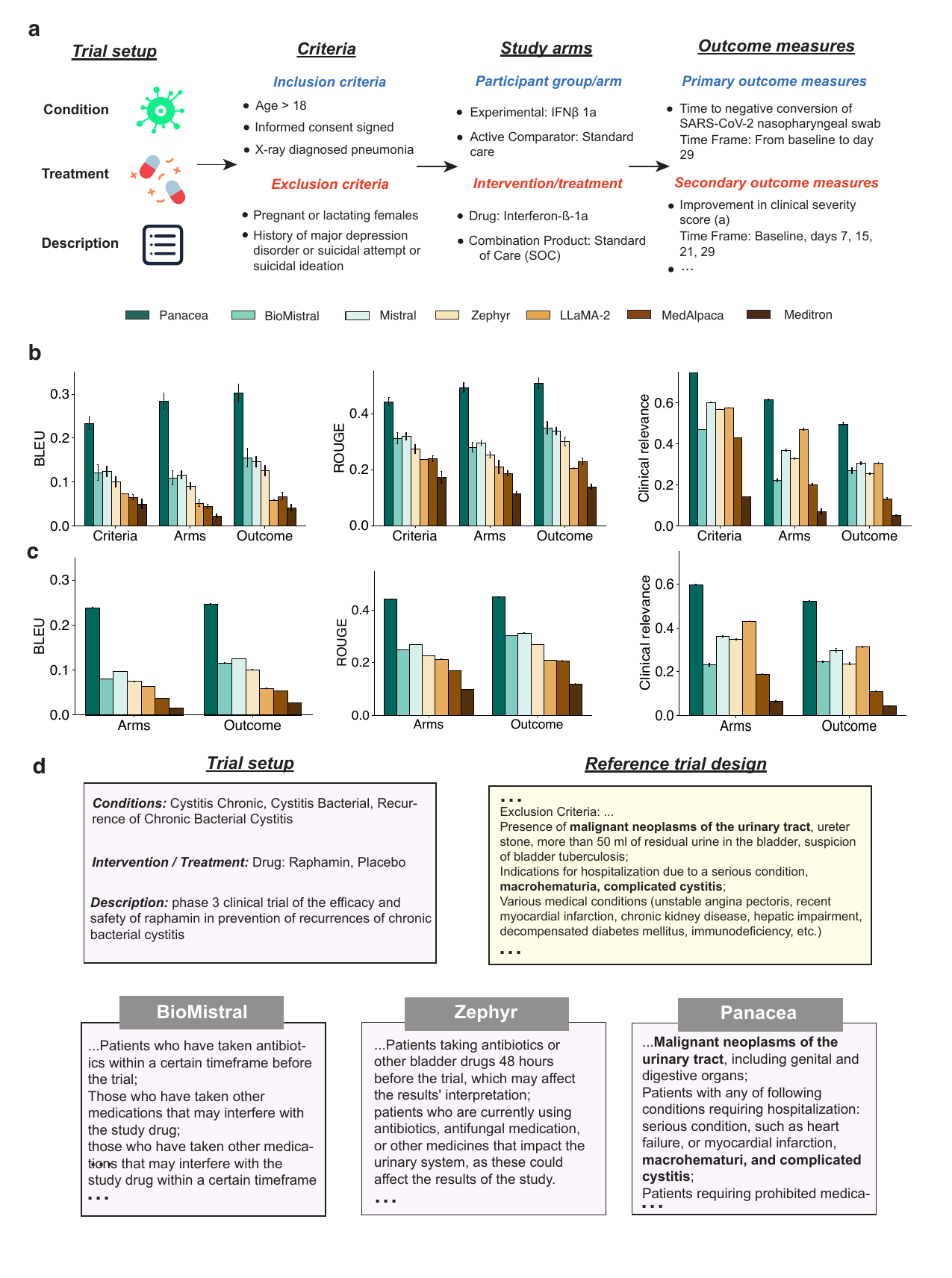} 
    \caption{\textbf{Evaluation on clinical trial design.} \textbf{a, } Problem setting of clinical trial design, which aims to generate criteria, study arms, and outcome measures. Criteria are used as input to generate study arms. Criteria and study arms are used to generate the outcome measures. 
\textbf{b, } Evaluation on trial design in terms of BLEU, ROUGE, and clinical relevance, where the reference design in the previous step is given as the input to the next step. \textbf{c, } Evaluation on trial design in terms of BLEU, ROUGE, and clinical relevance, where the generated design in the previous step is given as the input to the next step. \textbf{d, } A case study comparing criteria generation by different methods. \ourmethod can generate criteria that match the reference trial design.
}
    \label{fig:2design}
\end{figure}

The first step toward a successful trial execution is designing a detailed trial protocol synopsis.  We evaluated \ourmethod on three tasks in trial design (See examples in \textbf{Fig. \ref{fig:2design}a}): {\bf Criteria design} defines the eligibility criteria (i.e., the inclusion and exclusion criteria) for patient recruitment; {\bf Study arm design} outlines the different treatment arms that will be applied to different patient subgroups;  {\bf Outcome measures design} specifies the metrics that are used to assess the trial success. We formulated these three tasks as a conditional text generation problem, which takes conditions, treatments, and the design of previous steps (e.g., reference criteria are used to generate study arms) as inputs to generate specific design text.

Because trials are described in plain text, we first exploited standard natural language processing metrics BLEU and ROUGE to evaluate the lexical similarity. We found that \ourmethod attained the best performance on all three clinical trial design tasks in terms of BLEU and ROUGE (\textbf{Fig. \ref{fig:2design}b}). First, we observed that \ourmethod substantially outperformed general-domain models, including our base model Mistral \cite{jiang2023mistral}, confirming the benefit of fine-tuning using clinical trial-related data. Second, we found that \ourmethod improved the study arm design more than the other two tasks. Compared to criteria and outcome measures, study arm descriptions are more customized according to the disease and the treatment. The larger improvement of \ourmethod on study arms design demonstrates \ourmethod's strong generalization ability. Finally, BioMistral \cite{labrak2024biomistral}, which is fine-tuned on general biomedical data, also outperformed Mistral, further demonstrating the value of domain-specific data. Nevertheless, \ourmethod still outperformed BioMistral by fine-tuning using our clinical trial-specific data \ourdataalign and \ourdatainstr, suggesting that data with improving domain specificity leads to better performance. 

Lexical similarity metrics are widely used to evaluate text generation problems, but might not be clinically specific enough to evaluate the generations by \ourmethod. Recently, LLMs have been used to evaluate the generated text by exploiting their strong ability in text understanding. Here, we exploit Claude \cite{anthropic2024claude} to evaluate these three tasks by asking the model whether the generated task is clinically relevant (see \textbf{Methods}, \textbf{Supplementary Figures \ref{supfig:clinical_acc_criteria}\textbf{-}\ref{supfig:clinical_acc_outcome_measures}}). We found that \ourmethod outperforms all methods on criteria and study arms design, demonstrating the high quality of generation by \ourmethod (\textbf{Fig. \ref{fig:2design}b}). 
 
Moreover, we examined a De Novo generation setting, using the generated output in the previous step as the input for the next step. For example, we used the generated criteria instead of the reference criteria as the input for generating study arms. De Novo generation frees users from providing any descriptions for the trial. We found that the performance of all methods dropped in this setting compared to the setting that utilizes reference input (\textbf{Fig. \ref{fig:2design}c}). Nevertheless, our method still outperforms all existing methods by a large margin, indicating its superior performance on this De Novo trial design. We further compared the generated text by three methods with the ground truth text on criteria design, where only \ourmethod can generate the correct criteria (\textbf{Fig. \ref{fig:2design}d}). Collectively, the promising performance of \ourmethod demonstrates its potential to automate clinical trial design.

\subsection*{Accurate patient-trial matching}
\begin{figure}[!htbp]
    \centering
    \includegraphics[width=1.0\linewidth]{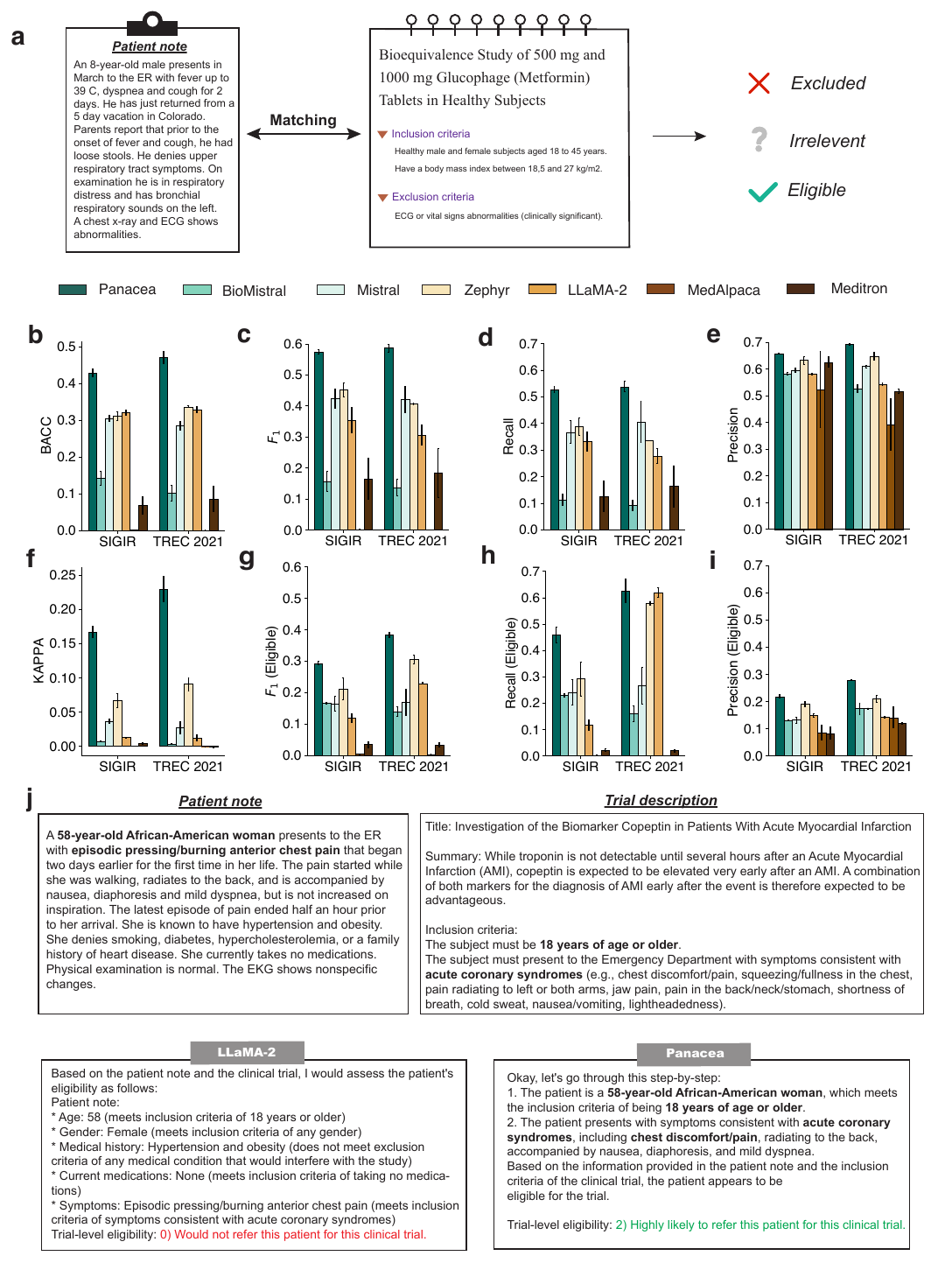} 
\caption{\textbf{Evaluation on patient-trial matching.} \textbf{a, } Problem setting of patient-trial matching, which classifies a patient into three categories based on the patient note and the trial description. \textbf{b-f} Comparison on two patient-trial matching datasets SIGIR and TREC 2021 in terms of balanced accuracy (BACC) (\textbf{b}), Cohen's KAPPA (\textbf{c}), recall (\textbf{d}), precision (\textbf{e}), and F1 (\textbf{f}). \textbf{g-i} Comparison on classifying patients into eligible and ineligible in terms of F1 (\textbf{g}), precision (\textbf{h}), and recall (\textbf{i}). \textbf{j,} A case study illustrating how \ourmethod successfully classifies the patient into eligible by examining each criterion.}
    \label{fig:3matching}
\end{figure}

We next evaluate the performance of \ourmethod on patient-trial matching. Given a patient note and a trial description, we aim to determine whether this patient is eligible for the trial by formulating this problem as a three-class classification task: eligible, excluded, or irrelevant (\textbf{Fig. \ref{fig:3matching}a}). 

We first evaluated our method on the TREC 2021 dataset \cite{roberts2021overview}, which consists of a training set and a test set. We used the training set to construct instructions in \ourdataalign, and then assessed the performance of \ourmethod on the test set. We found that \ourmethod outperformed all comparison approaches in terms of balanced accuracy (BACC), Cohen's KAPPA score, Recall, Precision, and F1, indicating the effectiveness of using \ourdatainstr to fine-tune the model (\textbf{Fig. \ref{fig:3matching}b-f}). To investigate the generalizability of our method, we further tested our method on the SIGIR dataset \cite{koopman2016test} where the entire dataset is used as the test set. We found that our method again attained the best performance on all three metrics, demonstrating the strong generalizability of our method.

As the eligible class is crucial for patient-trial recruitment, we further examined a binary classification setting. In this setting, we grouped "excluded" and "irrelevant" into one category, and "eligible" into the other in order to determine whether a patient is eligible for a trial. Our method outperformed all comparison approaches in terms of F1, precision, and recall, indicating its applicability to real-world trial recruitment (\textbf{Fig. \ref{fig:3matching}g-i}). Finally, we used a case study to illustrate how our method successfully classified a patient as eligible by examining each criterion and coming to a conclusion based on their criteria (\textbf{Fig. \ref{fig:3matching}j}). In contrast, LLaMA-2 \cite{touvron2023llama} made an incorrect conclusion by hallucinating an exclusion criterion not stated in the trial description.

\section*{Discussion}
In this paper, we introduce a specialized foundation model called \ourmethod for use in clinical trials. We tested \ourmethod in eight different clinical trial tasks, including trial design, patient-trial matching, trial search, and trial summarization. In comparison to other general domain foundation models and biomedical foundation models, \ourmethod demonstrated state-of-the-art performance across all eight tasks. We believe that the impressive performance of \ourmethod can be attributed to the fine-tuning process using \ourdataalign and \ourdatainstr. \ourdataalign comprises a large collection of trial documents and papers from various areas, allowing \ourmethod to be applied to different conditions and treatments. Meanwhile, \ourdatainstr contains 200,866 instructions curated from existing databases, effectively guiding \ourmethod in each task. Furthermore, we have developed a clinical trial benchmark \ourbenchmark and a language model-based metric for evaluating trial summarization. Together, these resources offer an end-to-end solution for AI-based clinical trial development.

The rapid development of large language models (LLMs) has enabled their potential as foundational models for medical tasks~\cite{moor2023foundation}. Current efforts predominantly follow two strategies: fine-tuning general domain LLMs with medical domain datasets~\cite{DBLP:journals/bib/LuoSXQZPL22,singhal2023large,chen2023meditron}, and instructing a general domain LLM with a description of the target tasks and showing example inputs and outputs (referred to as ``prompting'')~\cite{van2024adapted,tayebi2024large,DBLP:journals/corr/abs-2311-16452}. The MedPaLM model is a prime example of the first approach, illustrating how fine-tuning a general domain model on medical datasets can markedly enhance its ability to answer medical questions~\cite{singhal2023large}. This success has inspired further research into fine-tuning LLMs for specific clinical trial tasks, such as generating eligibility criteria~\cite{DBLP:conf/emnlp/0010X023}. Moreover, it has been demonstrated that generalist LLMs can be effectively adapted to medical tasks through strategic prompting~\cite{DBLP:journals/corr/abs-2311-16452}. In the direction of prompting, TrialGPT showcased that GPT-4 can be adapted to predict patient eligibility for clinical trials through prompting~\cite{DBLP:journals/corr/abs-2307-15051}. However, these approaches either do not address clinical trial tasks or focus on individual clinical trial-related tasks. In contrast, \ourmethod outlines a comprehensive range of clinical trial tasks suitable for AI assistance, establishing the first versatile foundational model specifically designed for clinical trial applications.

This study has several limitations that we would like to address in the future. First, despite being fine-tuned on clinical trial instruction datasets, LLMs may still produce biased or low-quality outputs. Enhancing model alignment such as reinforcement learning from human feedback~\cite{ouyang2022training} is crucial future work before \ourmethod can be deployed in production settings. Second, for high-stakes applications such as clinical trials, it is essential to detect and regulate LLM hallucinations, which can occur, particularly in areas not well-covered by the LLM training data. It is worth exploring to enable LLMs to either reject an answer~\cite{lin2023generating} or utilize external knowledge bases to correct its outputs~\cite{semnani2023wikichat}. Third, continually updating the model's knowledge is vital for maintaining relevance and accuracy in a rapidly evolving medical landscape. Therefore, it is worth exploring efficient knowledge updating techniques for \ourmethod~\cite{hu2021lora} or enhancing it with retrieval-augmented generation~\cite{lewis2020retrieval}. Fourth, although \ourmethod demonstrates significant improvements across various benchmark datasets, there is a need to develop more evaluation metrics to comprehensively assess LLM performance in more clinical trial tasks. Additionally, conducting user studies could further demonstrate the benefits of \ourmethod in assisting experts with clinical development projects.

% Fourth, LLM scaling laws suggest that scaling up the parameter size of \ourmethod could probably push the boundary of AI's capability in clinical trials~\cite{wei2022emergent}.

% There are at least three limitations of \ourmethod we would like to address in the future. First, evaluation is hard, need to design better metric, similar to our effort in the trial summarzation. Second, add conversation ability to engage with users? Third, explore how to use it togewther with drug discovery model, disease prediction model, etc. 
\newpage
\section*{Method}
\subsection*{Creating \ourdataalign dataset}
\noindent \textbf{Data collection} We first collected trial documents (English version) from 14 sources, as shown in \textbf{Supplementary Table \ref{tab:stat_datasets}}. Each clinical trial data consists of various parts that encapsulate the essence of the study. For instance, the “Study Overview” provides a general summary and a detailed description of the trial, along with its official title and the health conditions being targeted. The “Intervention/Treatment” section describes the medical approach or therapy being tested. The “Eligibility Criteria” outlines who can participate, detailing the eligibility requirements, age, and sex specifications, and whether healthy volunteers are accepted. The “Study Plan” delves into the methodology, explaining the design of the study, the types of interventions and arms involved, and the outcomes being measured, both primary and secondary. This structured approach ensures a comprehensive understanding of the trial's scope, methodology, and intended outcomes. We then collected trial papers in two databases, i.e., Embase and PubMed, from Cochrane Library's trial section \cite{cochranecentral}. These papers provide a rich foundation of medical knowledge and evidence-based findings beneficial to the model's learning. 

\noindent \textbf{Filtering} For trial documents, we further conduct intra- and inter-source de-duplication and then remove the personally identifiable information (PII), finally obtaining 793k trial document data. Further, to avoid information leakage, we selected documents with registration dates before 2023-01-01 as the training corpus. The remaining is used for test data curation. For trial papers, we de-duplicated all the papers and the final 1.11M trial paper corpus consists of abstracts of all the papers and full text of 97k papers from PubMed Central (PMC). Similarly, to avoid information leakage, we choose papers published before 2023-01-01, which ensures the dates of related clinical trials of the selected papers are definitely before 2023-01-01.

\noindent \textbf{Document/paper structure organization} For trial documents, we follow the format shown in clinicaltrial.gov \cite{clinicaltrialsgov} to organize all the corpus for alignment. Each trial document is arranged into a markdown format passage. For trial documents from clinicaltrial.gov, each document contains section (1) “Public Title”; (2) “Study Overview” covering subsections “Brief Summary”, “Detailed Description”, “Official Title”, “Conditions” and “Intervention/Treatment”; (3) Participation Criteria, including subsections “Eligibility Criteria”, “Ages Eligibility for Study”, “Sexes Eligibility for Study” and “Accepts Healthy Volunteers”; (4) “Study Plan”, including subsection “How is the study designed?” that contains “Design Details” and “Arms and Interventions”, subsection “What is the study measuring?” containing primary and secondary outcome measures; (5)Terms related to the study. For trial documents from other sources, each document contains “Public Title”, “Scientific Title”, “Study Type”, “Study Design”, “Intervention”, “Inclusion Criteria”, “Exclusion Criteria”, “Primary Outcome Measures” and “Secondary Outcome Measures”. For trial paper data, each paper contains “Title”, “Abstract” and full text (if any).

\subsection*{Creating \ourdatainstr dataset}
\label{subsec:inst}
The aim of constructing \ourdatainstr is to provide \ourmethod with the ability to follow human instructions, especially in clinical trial domains. 

\noindent \textbf{Trial search} Trial search includes query generation and query expansion. To construct instruction data for query generation, we leverage GPT-3.5 to generate 2,161 samples for training and 925 for the test. Specifically, we first manually construct 20 seed data about query generation customized for clinicaltrial.gov database API, and then leverage GPT-3.5 to generate the data. We will remove data similar to the original data and add them to the seed dataset to repeat the above process (see prompt in \textbf{Supplementary Figure \ref{supfig:prompt_query_gen_data_gen}}). In the final stage, we send requests with these generated data to the clinicaltrial.gov database and remove those without any search results. For query expansion data curation, we turn to the mesh terms section in clinicaltrial.gov documents. Each document contains synonymous mesh terms. We keep five terms for each document as input and the others as output. For example, the input mesh terms are Gastroenteritis, Gastrointestinal Diseases, Digestive System Diseases, Colonic Diseases, Intestinal Diseases, Pathologic Processes, while the output terms are Inflammatory Bowel Diseases, Ulcer, Anti-Bacterial Agents, and Vancomycin. We select documents before 2023-01-01 for training and after 2023-01-01 for test. We finally obtained 50k training data and 2,500 test data.

\noindent \textbf{Trial summarization} Trial summarization contains single-trial and multi-trial summarization. To curate single-trial summarization data, we leverage clinicaltrial.gov documents. Specifically, the brief summary section serves as the output and the other parts serve as the input. We finally have 5k training data (before 2023-01-01)  and 1k test data (after 2023-01-01). For the multi-trial summarization data curation, we derived our dataset from Cochrane dataset of systematic reviews \cite{wallace2021generating}, i.e., we only selected data pairs containing clinical trial papers. Specifically, each multi-trial summarization data contains a PMID set and a review paper. The review is a high-level conclusion from papers in the PMID set. The data curation process started with the matching between the PMID sets and all the trial paper PMIDs in \ourdataalign. We select those data pairs with at least three trial-related papers in the PMID set. We finally constructed 2,029 samples for training and 252 for test, derived from the Cochrane dataset’s training and validation sets due to the missing test labels in the original Cochrane dataset. 

\noindent \textbf{Trial design} We construct multi-turn conversation data for trial design due to the difficulty of one-turn design, even for frontier models like GPT-4 \cite{achiam2023gpt}. Such conversation format data are more realistic and benefit users to get more accurate designs as conversations progress. To construct these conversation data, we focus on trial documents in clinicaltrial.gov and adopt a two-stage strategy to construct the conversation data. For criteria design, we first input criteria and trial setup, which contains title, conditions, drugs, and phase, to ask GPT-3.5 to output the reasons for designing those criteria one by one. In the second stage, we input the criteria, and reasons generated in the first stage, and trial setup, to ask GPT-3.5 to construct multi-turn conversation data (see \textbf{Supplementary Figure \ref{supfig:prompt_gen_criteria_design}}). This can ensure that GPT-3.5 generated trial part data is actual. Likewise, for study arm design, we input study arms, criteria, and trial setup. In the second stage, we collect the generated conversation data given the study arms, reasons, criteria, and trial setup (see \textbf{Supplementary Figure \ref{supfig:prompt_gen_study_arm_design}}). For outcome measures, the input in the first stage is outcome measures, study arms, criteria, and trial setup, while the input in the second stage is outcome measures, reasons, study arms, criteria, and trial setup (see \textbf{Supplementary Figure \ref{supfig:prompt_gen_soutcome_measure_design}}). We use trial documents from clinicaltrial.gov to construct these data, before 2023-01-01 for training and after 2023-01-01 for testing. We finally obtained 35,951 and 549 for the criteria design’s training and test set, 53,548 and 549 for the study arm design, and 44,809 and 549 for the outcome measure design.

\noindent \textbf{Patient-trial matching} We converted existing representative patient-trial matching datasets into instruction format, i.e., SIGIR \cite{koopman2016test} and TREC 2021 \cite{roberts2021overview} cohorts. Each instruction data of patient-trial matching follows the structure: “Instruction”, “One-shot demonstration”, “Input patient notes”, “Input Criteria” and “Output trial-level eligibility”, as illustrated in \textbf{Supplementary Figure \ref{supfig:patient_trial_matching}}. We split the TREC 2021 into the training  (28,406 samples) and test sets (7,424 samples), and all SIGIR data serves as the test set (3,869 samples). Specifically, the patient-criteria pairs of 80\% of patients in TREC 2021 formed into the training set, while those pairs of the remaining 20\% of patients in TREC 2021 are test data. For evaluation, we trained our \ourmethod on the training set derived from TREC 2021 and evaluated on the test set of TREC 2021 and all data in SIGIR.

\subsection*{Creating \ourbenchmark benchmark}

We built the first large-scale benchmark \ourbenchmark, including eight tasks in clinical trials. The training and test data constructed in the previous section are viewed as the benchmark data. We evaluated the models on \ourbenchmark to assess each model's performance across different clinical trial tasks.

% Metrics

\subsection*{Details of \ourmethod model}
In this section, we detail the techniques in \ourmethod, including the alignment and instruction finetuning steps.

\noindent \textbf{Alignment} We built on the Mistral-7B-Base model \cite{jiang2023mistral} in this study. After parameter initialization, \ourmethod was trained on the 1.8M \ourdataalign data. We trained the model using the AdamW optimizer \cite{DBLP:conf/iclr/LoshchilovH19} with a batch size 512 for one epoch. We adopted a cosine learning rate scheduler with a peak learning rate $2 \times 10^{-6}$ and 10\% warm-up steps. We set max sequence length as 8192 tokens. To improve training speed and optimize the memory, we adopted DeepSpeed ZeRO-3 \cite{rajbhandari2020zero} and FlashAttention-2 \cite{dao2023flashattention} strategies. After the alignment process, we obtain the \ourmethod-Base model. During the alignment step, \ourmethod was trained on 4 Nvidia A100 80G for four days.

\noindent \textbf{Instruction tuning} We further finetuned \ourmethod-Base on the \ourdatainstr datasets, leading to the \ourmethod model. We trained our \ourmethod for one epoch with a batch size 256. Similar to the alignment step, we also leveraged a cosine learning rate scheduler with a peak learning rate as $2\times10^{-5}$ and 10\% warm-up steps. The max sequence length is set as 2048. Deep ZeRO-3 and FlashAttention-2 techniques are also adopted in the instruction tuning phase.

\subsection*{Details of experiments on trial search}
% query generation -> jsonformer
% query expansion 
In the trial search experiments, we focused on optimizing \ourmethod for two tasks: query generation and query expansion (see \textbf{Supplementary Figure \ref{supfig:trial_search}}). These two tasks are pivotal for enhancing the efficiency and precision of searches within large clinical trial databases.

Query generation in this context essentially functions as a Named Entity Recognition (NER) task where the model identifies and categorizes key pieces of information from the trial descriptions relevant to user queries. To facilitate the generation of structured queries in a JSON format, we employed a specialized tool called JsonFormer \cite{jsonformer}. This tool is instrumental in guiding the model to generate content for each key in the JSON structure sequentially.

Once the JSON format is generated, it is automatically converted into a Search Expression using a rule-based system. The conversion rules are straightforward: within the same key, terms are combined using the OR operator, and between different keys, the terms are combined using the AND operator. This structured approach ensures that the generated queries are precise and align well with the syntactical requirements of the search engines used in clinical trial databases.

For the query expansion task, this process enhances the original query by adding semantically related terms, thereby broadening the search scope to include relevant trials that may not use the exact phrasing of the original query terms. \ourmethod was trained to suggest additional keywords based on the initial input terms. The model learned to recognize and predict related terms that could be associated with the initial query, expanding the search breadth effectively.

\subsection*{Details of experiments on trial summarization}
The experiments on trial summarization were designed to test \ourmethod's capabilities in condensing complex clinical trial information into succinct summaries. This component of our research focused on two specific tasks: single-trial summarization and multi-trial summarization (see \textbf{Supplementary Figure \ref{supfig:trial_summarization}}).

To evaluate summarization tasks, we propose a novel metric based on Claude 3. We use Claude 3 to decide whether the model-generated summarization and the ground truth summarization studied the same problem and made the same conclusion, following prompts in \textbf{Supplementary Figure \ref{supfig:summary_eval_gpt} and \ref{supfig:summary_eval_gpt_multi}}. Specifically, Claude 3 directly outputs the goal alignment results for each test sample. For conclusion consistency, we first use Claude to evaluate model-generated summaries and ground truth summaries, respectively. Then, we calculate the matching accuracy between the model-generated summarization and ground truth summarization.

\subsection*{Details of experiments on clinical trial design}
% after 3 steps, input the conversation history -> output next. for next turn, we use the groudtruth history, in case of deviate far.
In our experimental setup for evaluating the \ourmethod model's capabilities in clinical trial design, we utilized a multi-turn conversation format for the test data. This format consists of sequential (user, chatbot) pairs, reflecting a realistic interaction scenario where the model, acting as a chatbot, responds to user queries about designing a trial. The initial three rounds usually provide essential background information related to the trial design, such as the trial's objectives, target population, and key endpoints. These initial conversations set the stage for the more complex interactions that follow. Starting from the fourth round of conversation, the model is tasked with predicting the chatbot's responses based on the cumulative conversation history, which tests the model's ability to maintain context and continuity over successive interactions. 

To ensure the reliability of the experimental results and prevent the propagation of errors through the conversation chain, a teaching forcing strategy was implemented: regardless of the model's output in any given round, the subsequent round's input incorporates the groundtruth from the previous rounds rather than the model-generated responses. This method allows the model to be evaluated on its ability to adhere closely to a scientifically valid trial design path without being influenced by potential errors in its previous outputs.

To assess the relevance between models' designed trials and ground truth, we employ Claude 3 to calculate clinical relevance. Specifically, we input each model's output and the ground truth into Claude 3 to determine the relevance of the information generated by the model compared to the ground truth. The inputs to Claude 3 for clinical relevance evaluation are detailed in \textbf{Supplementary Figures \ref{supfig:clinical_acc_criteria}, \ref{supfig:clinical_acc_study_arms}, and \ref{supfig:clinical_acc_outcome_measures}}, respectively. When a model's outputs are relevant to the ground truth, Claude will output a 1; otherwise, it outputs a 0. We then calculate the clinical relevance using the following formula:

\begin{equation}
    \text{Clinical relevance} = \cfrac{\sum(\text{Relevance scores})}{N}
\end{equation}
Here, ``Relevance scores" refer to the series of 1s and 0s output by Claude 3 for each comparison between a model's output and the ground truth. $N$ is the total number of outputs evaluated. This proportion reflects the percentage of times the model's output was deemed clinically accurate relative to the ground truth, quantifying the frequency at which the model produces clinically relevant information.

\subsection*{Details of experiments on patient-trial matching}
% Use claude-3 for filtering. Difference between two datasets. 
In the patient-trial matching experiments, we employed a distinctive approach to training the \ourmethod model, focusing not on utilizing the entirety of the training data but rather on a selected subset. Initially, all available training data was subjected to a filtering process with Claude 3 Haiku. This involved predicting responses for each instance in the training set. Only those instances where Claude 3 Haiku's predictions were accurate were retained for further processing. The rationale was to ensure that the model was learning from correctly reasoned examples and that the training data was high quality. The responses generated by Claude 3 Haiku, which correctly matched the groundtruth data, were then used as the new training corpus for \ourmethod. This step was crucial because the standard training datasets for patient-trial matching typically include labels indicating eligible or excluded but lack a detailed reasoning process for these outcomes. By incorporating Claude 3 Haiku's responses, which involve step-by-step reasoning based on the input data, we injected reasoning capabilities into \ourmethod during the training process. Through this innovative training approach, \ourmethod showed superior performance in patient-trial matching tasks. The ability to reason and logically process eligibility criteria translated into higher accuracy and reliability in matching patients to appropriate trials. The evaluation prompt for patient-trial matching can be seen in \textbf{Supplementary Figure \ref{supfig:patient_trial_matching}}.

The patient-trial matching is a three-class classification task for both SIGIR and TREC2021 datasets. Three classes for SIGIR are: 0) Would not refer this patient for this clinical trial; 1) Would consider referring this patient to this clinical trial upon further investigation; and 2) Highly likely to refer this patient for this clinical trial, while TREC2021 has: 0) Excluded (patient meets inclusion criteria, but is excluded on the grounds of the trial's exclusion criteria); 1) Not relevant (patient does not have sufficient information to qualify for the trial); and 2) Eligible (patient meets inclusion criteria and exclusion criteria do not apply).

\section*{Code and data availability}
The \ourdataalign data for the alignment step, the \ourdatainstr data for the instruction tuning step, and the \ourbenchmark benchmark data are available at \url{https://figshare.com/articles/dataset/TrialAlign/25989403}, \url{https://doi.org/10.6084/m9.figshare.25990090.v1}, and \url{https://doi.org/10.6084/m9.figshare.25990075}, respectively. Panacea code is available at \url{https://github.com/linjc16/Panacea}.

\setcounter{table}{0}
\renewcommand*{\tablename}{Supplementary Table}

\setcounter{figure}{0}
\renewcommand*{\figurename}{Supplementary Figure}

\begin{figure}[!htbp]
    \centering
    \includegraphics[width=1.0\linewidth]{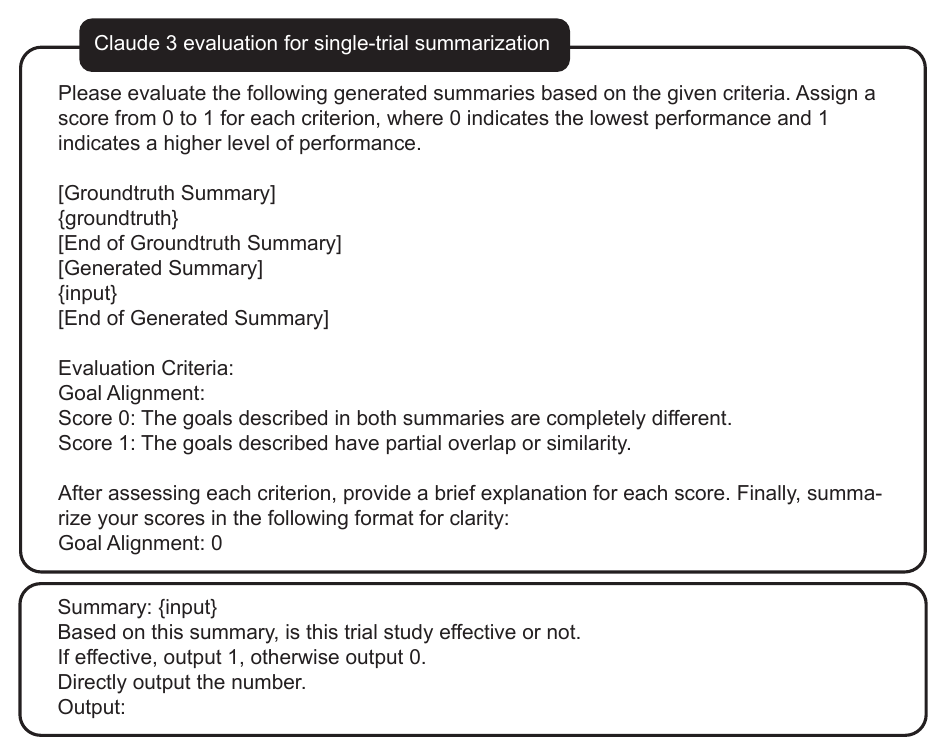} 
    \caption{Prompt for evaluation metrics on single-trial summarization.}
    \label{supfig:summary_eval_gpt}
\end{figure}

\begin{figure}[!htbp]
    \centering
    \includegraphics[width=1.0\linewidth]{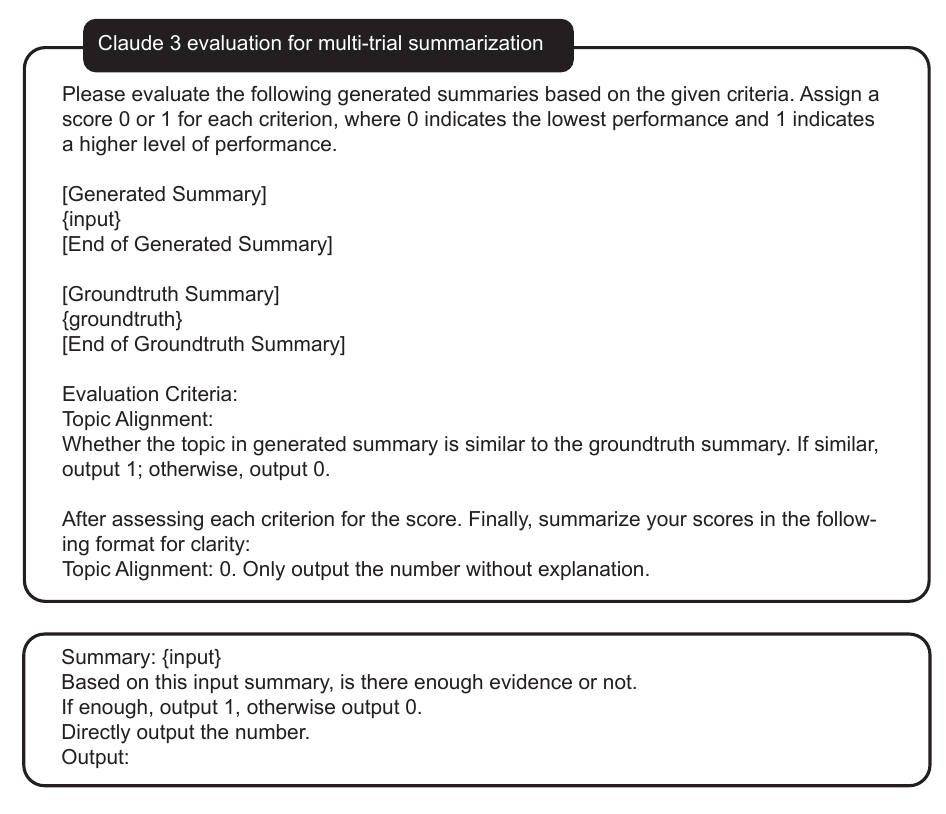} 
    \caption{Prompt for evaluation metrics on multi-trial summarization.}
    \label{supfig:summary_eval_gpt_multi}
\end{figure}

\begin{figure}[!htbp]
    \centering
    \includegraphics[width=1.0\linewidth]{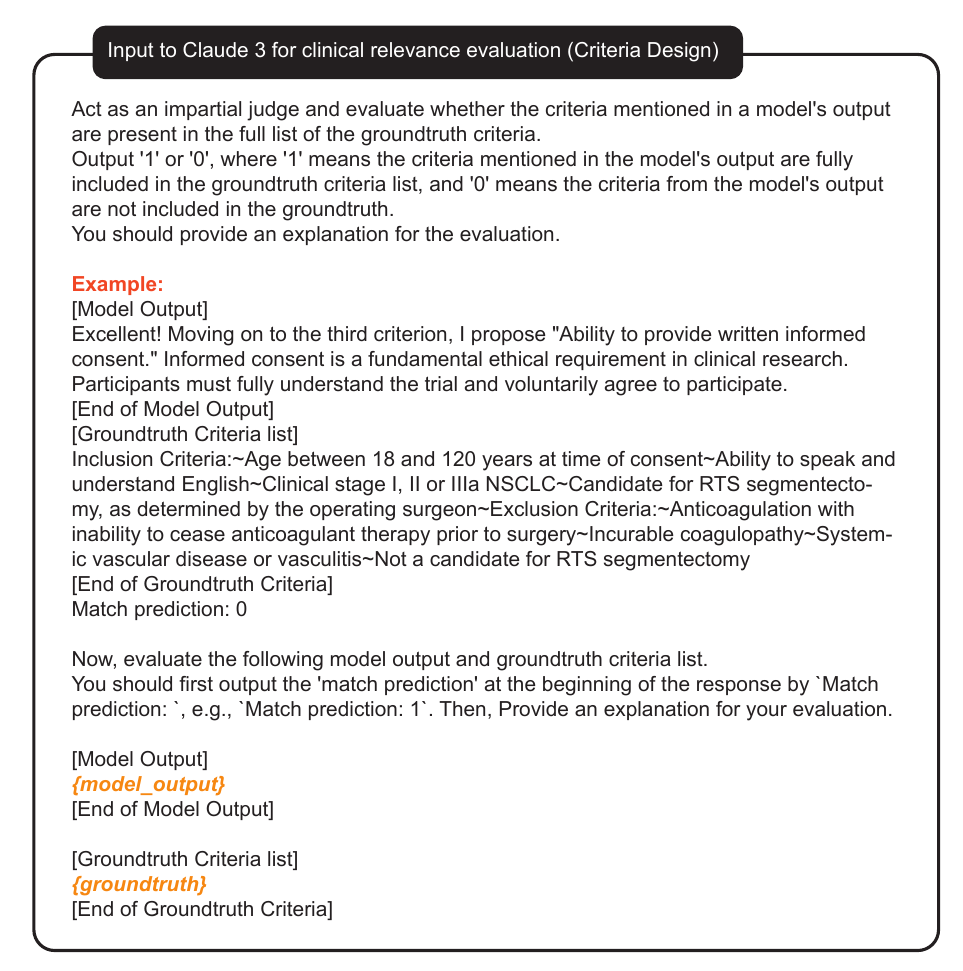} 
    \caption{Prompt used to calculate clinical relevance for criteria design.}
    \label{supfig:clinical_acc_criteria}
\end{figure}

\begin{figure}[!htbp]
    \centering
    \includegraphics[width=1.0\linewidth]{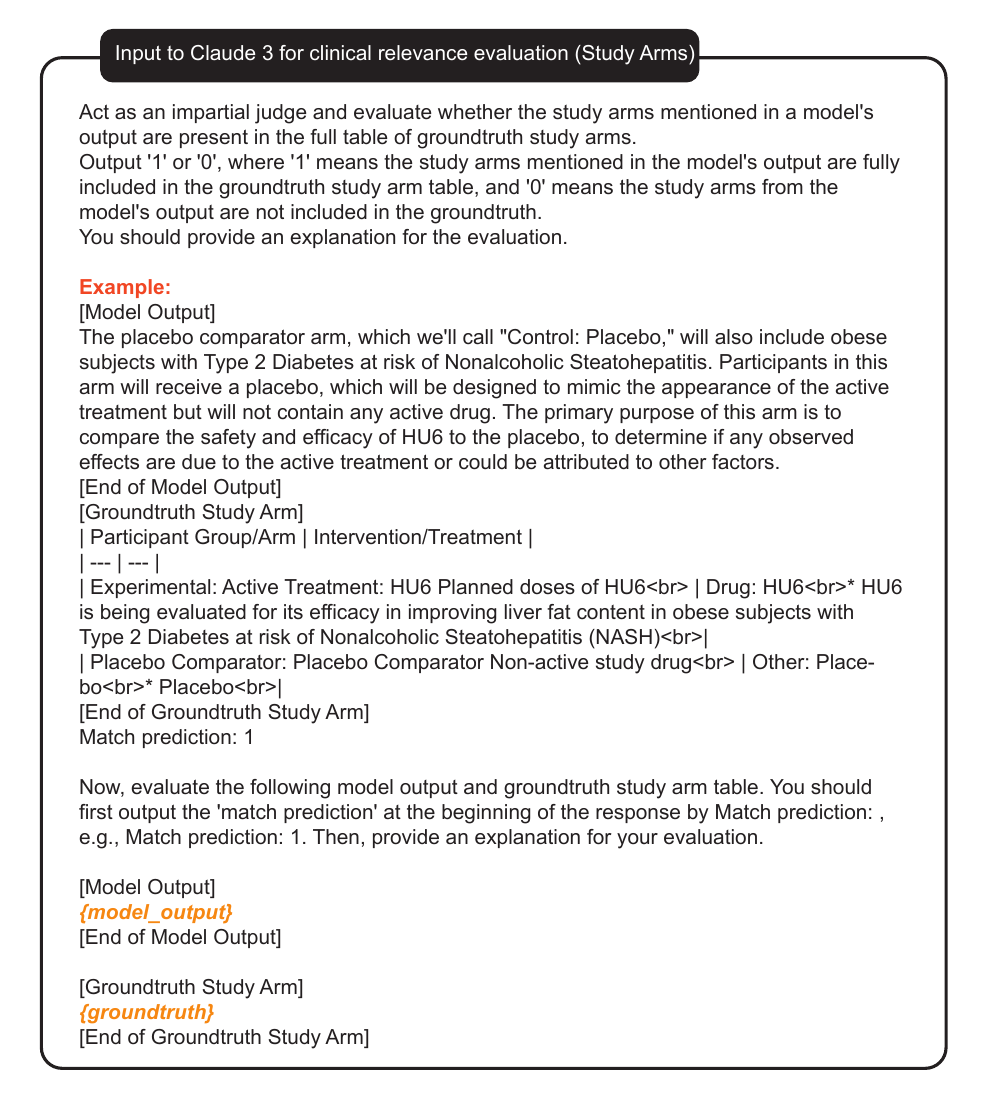} 
    \caption{Prompt used to calculate clinical relevance for study arms.}
    \label{supfig:clinical_acc_study_arms}
\end{figure}

\begin{figure}[!htbp]
    \centering
    \includegraphics[width=1.0\linewidth]{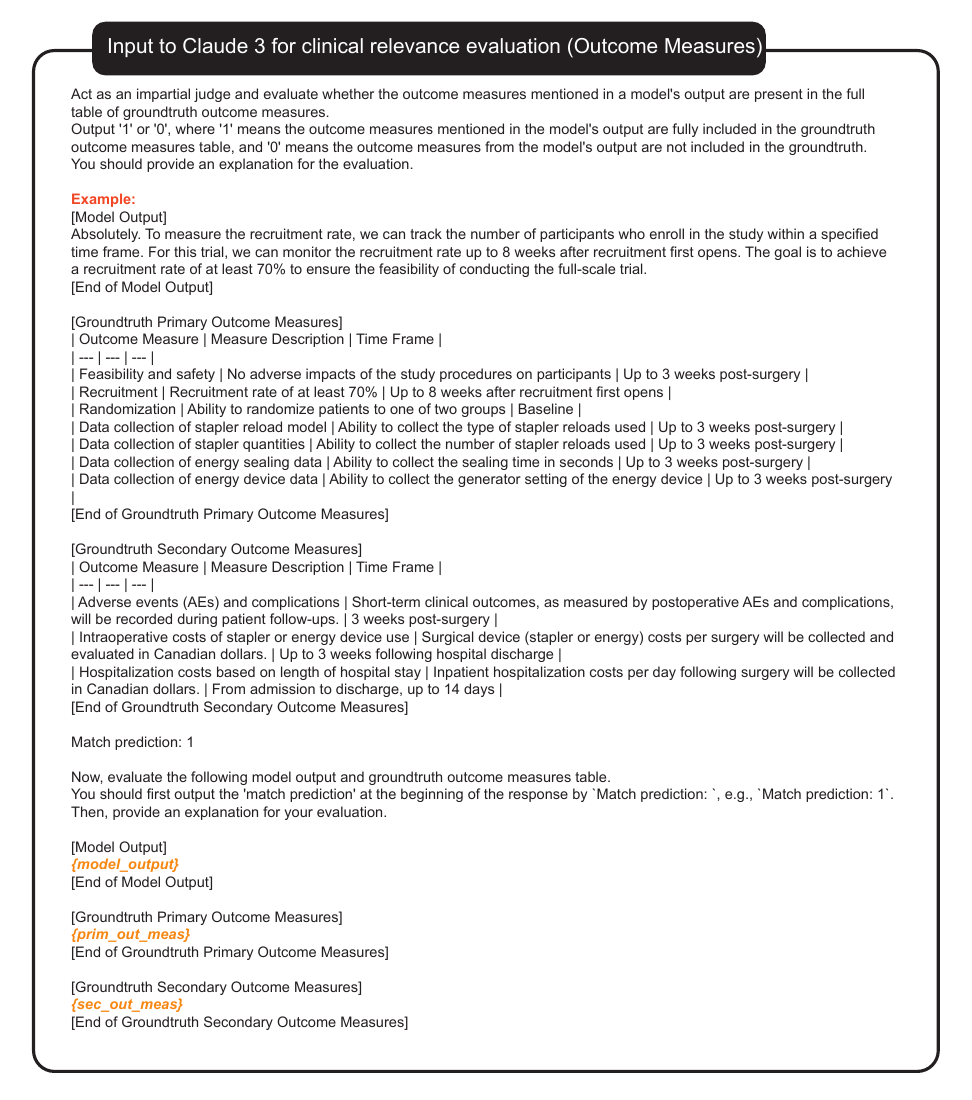} 
    \caption{Prompt used to calculate clinical relevance for outcome measures.}
    \label{supfig:clinical_acc_outcome_measures}
\end{figure}

\begin{figure}[!htbp]
    \centering
    \includegraphics[width=1.0\linewidth]{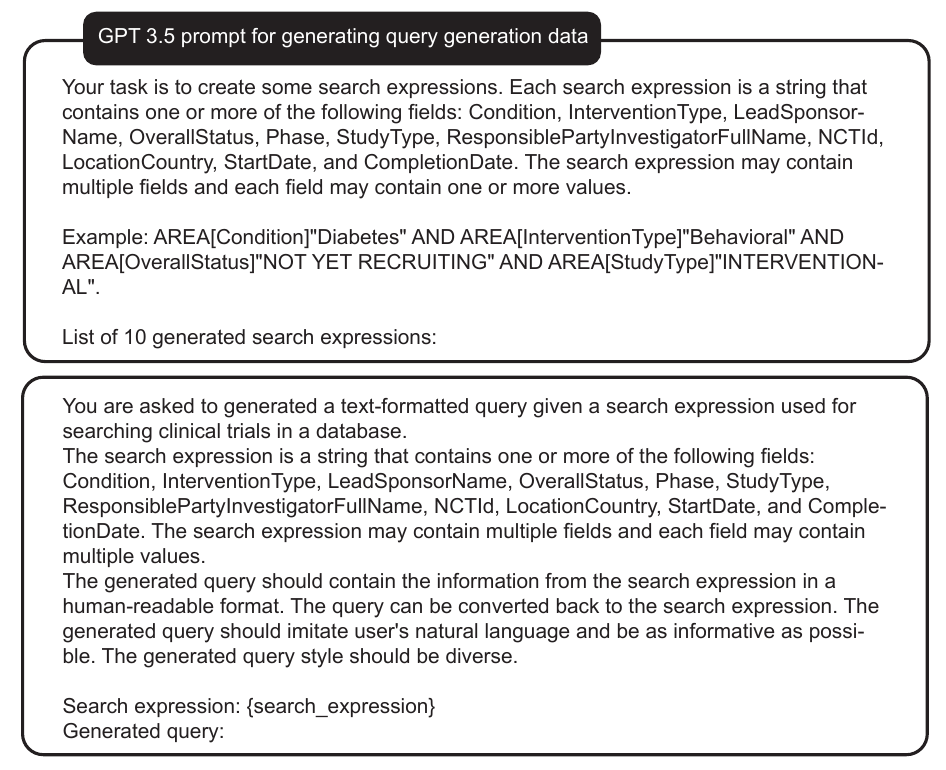} 
    \caption{Prompt used to construct query generation task data with GPT-3.5.}
    \label{supfig:prompt_query_gen_data_gen}
\end{figure}

\begin{figure}[!htbp]
    \centering
    \includegraphics[width=1.0\linewidth]{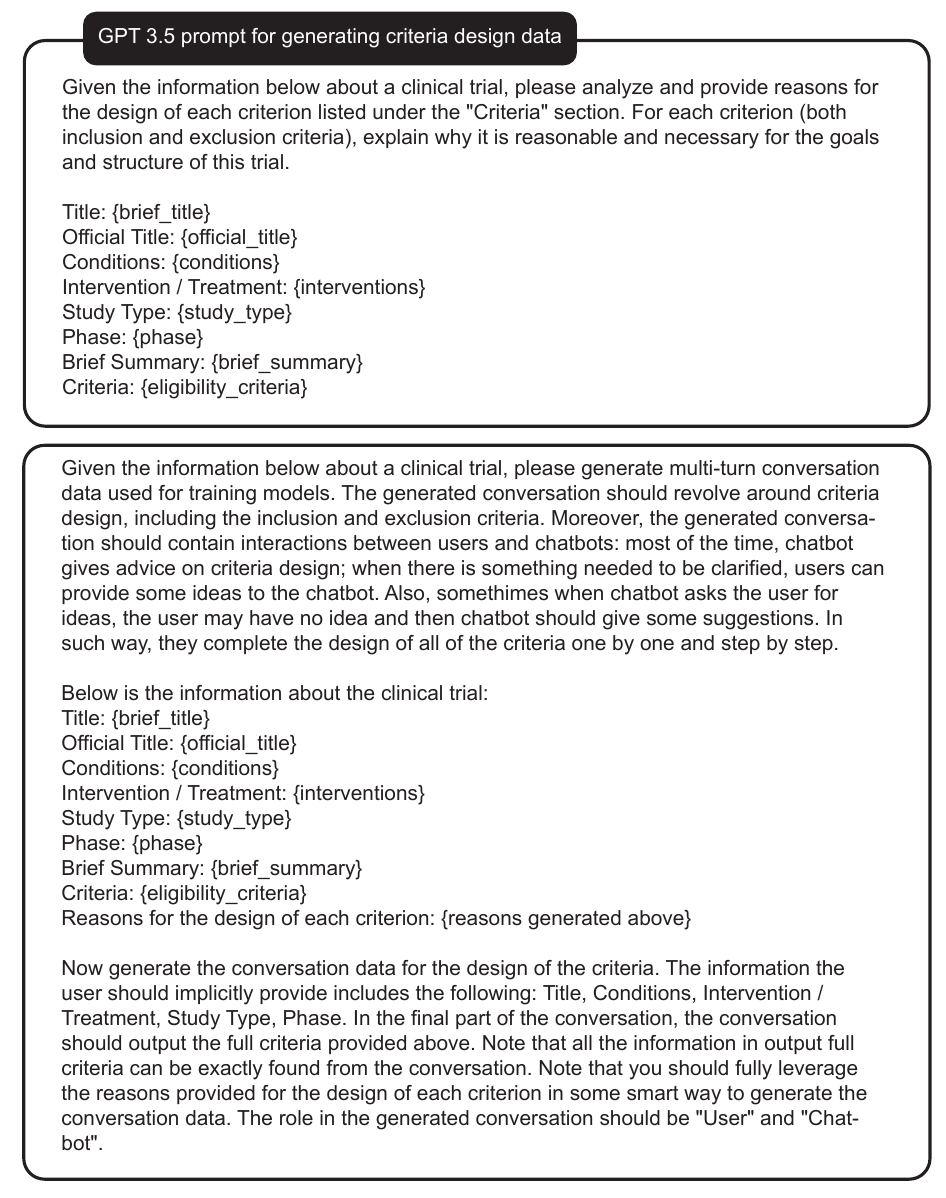} 
    \caption{Prompt for generating criteria design conversation data.}
    \label{supfig:prompt_gen_criteria_design}
\end{figure}

\begin{figure}[!htbp]
    \centering
    \includegraphics[width=1.0\linewidth]{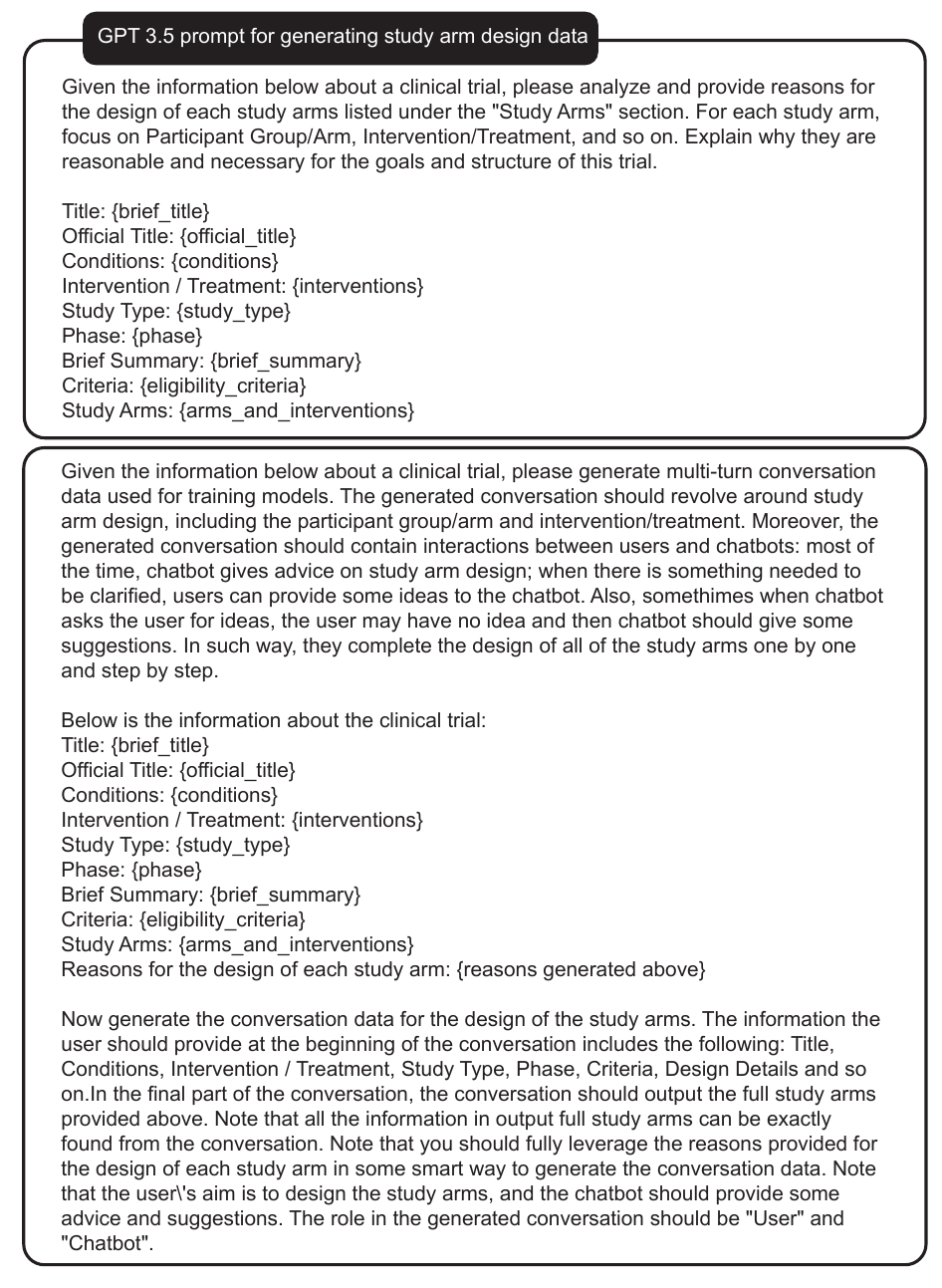} 
    \caption{Prompt for generating study arm design conversation data.}
    \label{supfig:prompt_gen_study_arm_design}
\end{figure}

\begin{figure}[!htbp]
    \centering
    \includegraphics[width=1.0\linewidth]{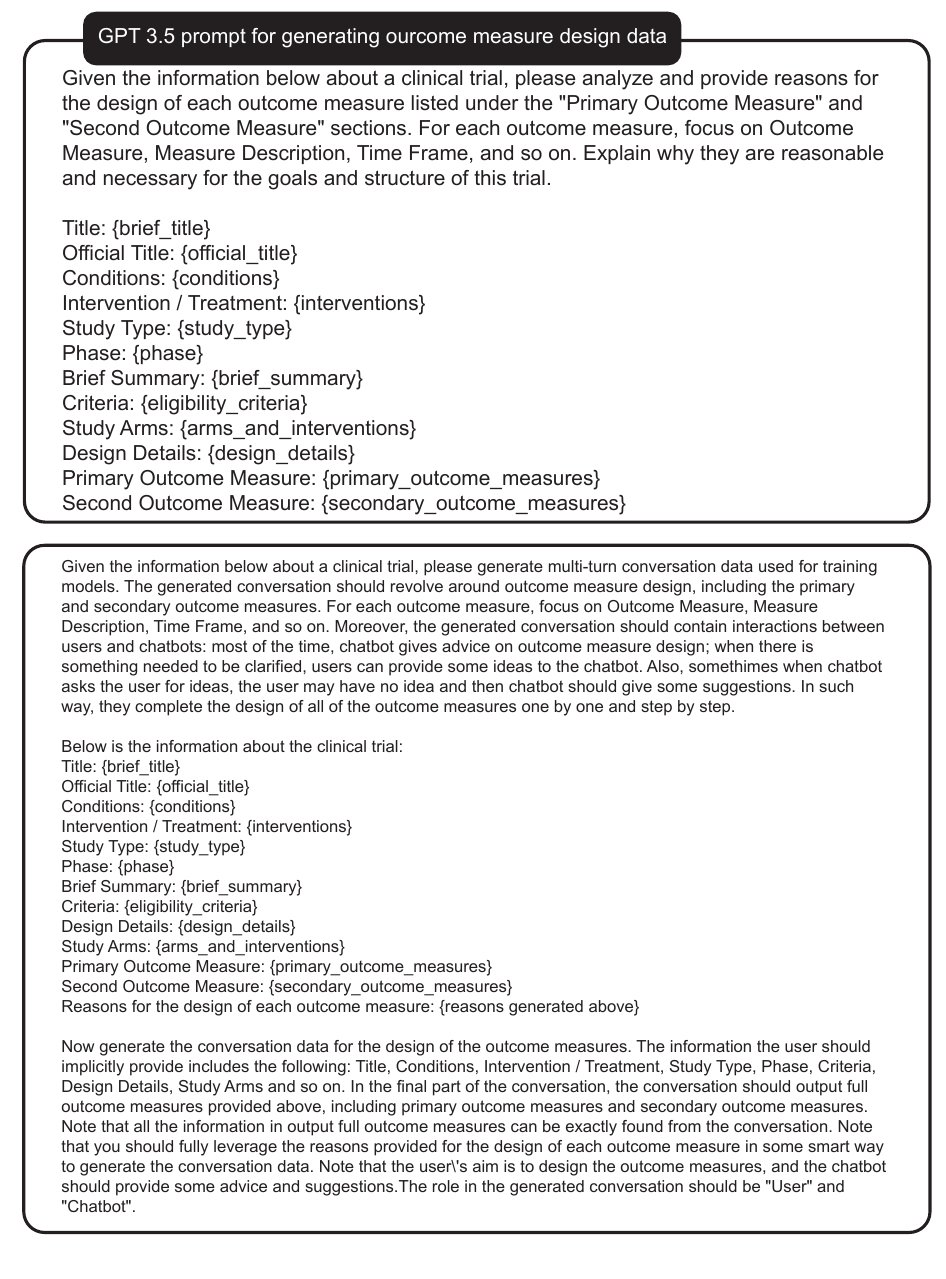} 
    \caption{Prompt for generating outcome measure design conversation data.}
    \label{supfig:prompt_gen_soutcome_measure_design}
\end{figure}

\begin{figure}[!htbp]
    \centering
    \includegraphics[width=1.0\linewidth]{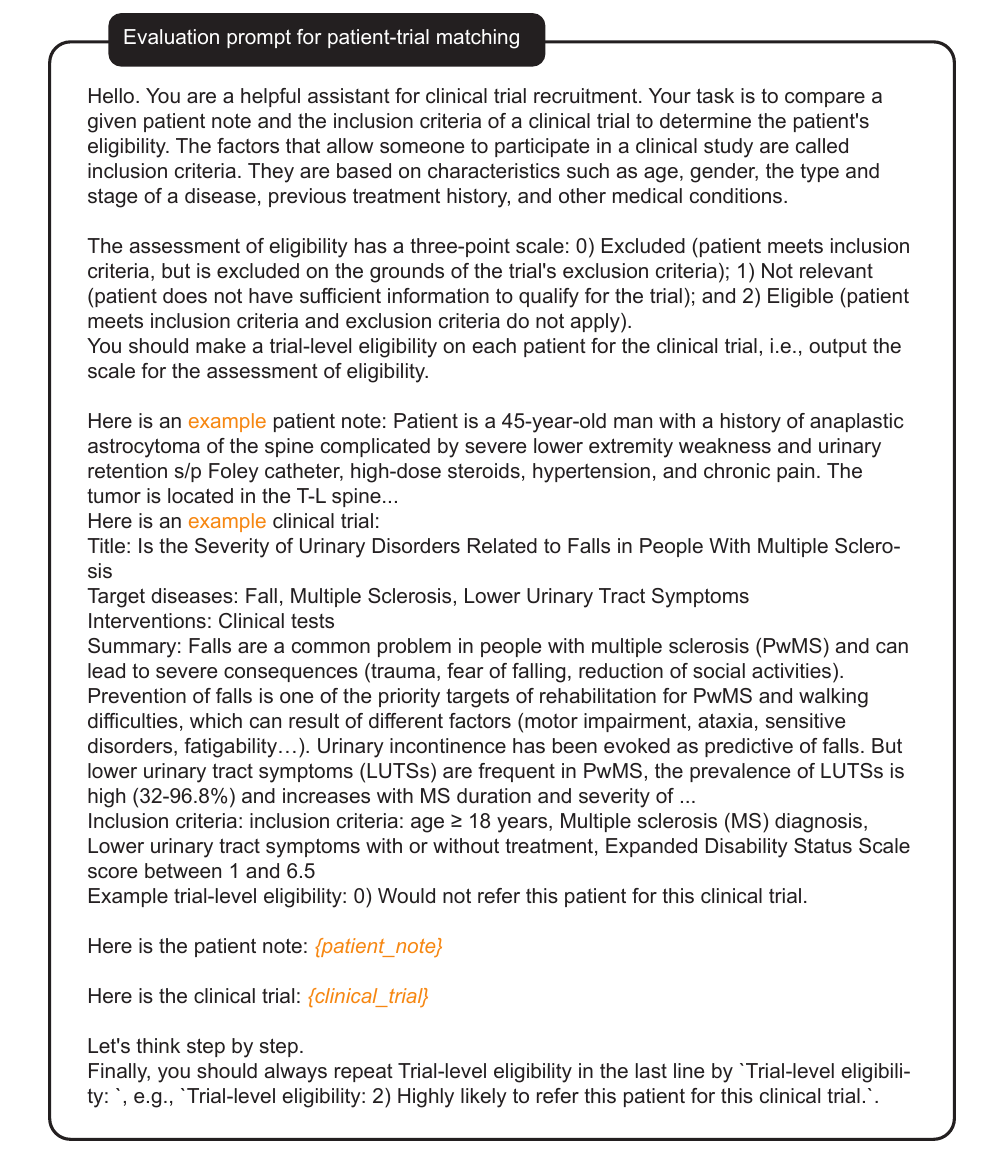} 
    \caption{Prompt for evaluation on patient-trial matching.}
    \label{supfig:patient_trial_matching}
\end{figure}

\begin{figure}[!htbp]
    \centering
    \includegraphics[width=1.0\linewidth]{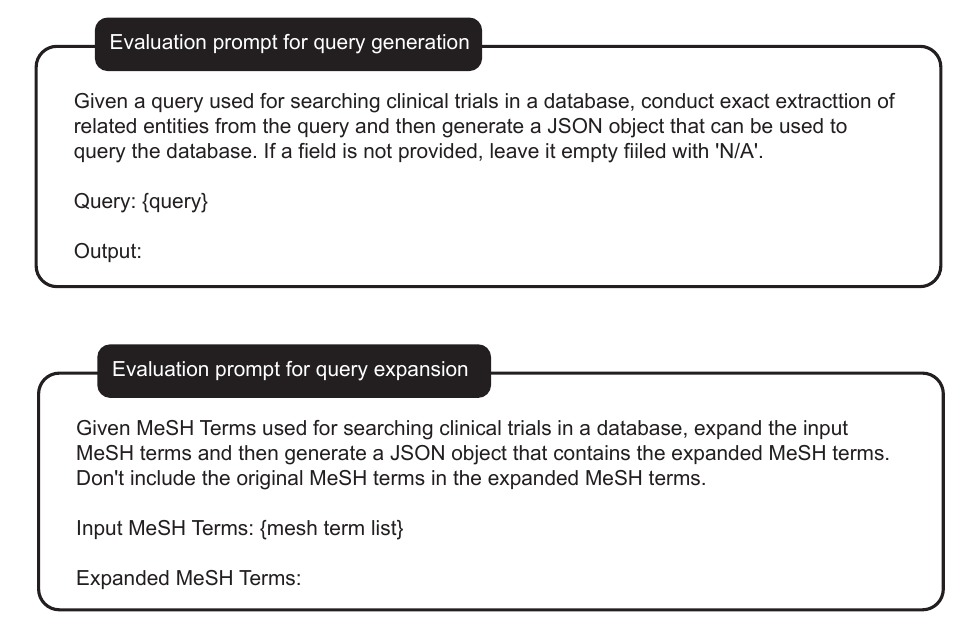} 
    \caption{Prompt for evaluation on trial search.}
    \label{supfig:trial_search}
\end{figure}

\begin{figure}[!htbp]
    \centering
    \includegraphics[width=1.0\linewidth]{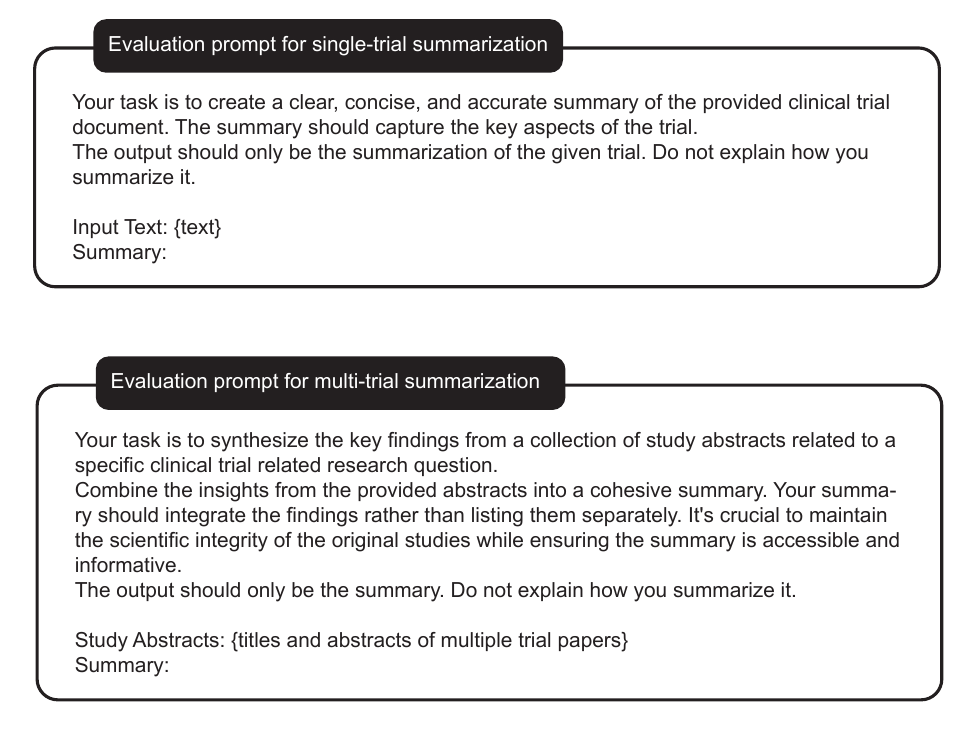} 
    \caption{Prompt for evaluation on trial summarization.}
    \label{supfig:trial_summarization}
\end{figure}

\begin{table}[ht]
\centering
\caption{Statistics of \ourdataalign. }
\label{tab:stat_datasets}
\resizebox{\textwidth}{!}{
\begin{tabular}{lccc}
\toprule
Source                 & \# Total   & \# Train ($<$ 2023) & \# Test ($\geq$ 2023) \\
\midrule
ClinicalTrials.gov \cite{clinicaltrialsgov}     & 467,944 & 432,676                 & 31,023                       \\
ChiCTR (China)   \cite{chictr}               & 76,186  & 65,181                  & 11,005                       \\
EUCTR (EU) \cite{euclinicaltrialsregister}       & 43,599  & 43,315                  & 284                         \\
JRCT (Japan)    \cite{japanrctportal}       & 64,650  & 60,645                  & 4,005                        \\
ANZCTR (Australian New Zealand) \cite{anzctr} & 24,657  & 23,374                  & 1,283                        \\
ISRCTN.org \cite{isrctnregistry}            & 24,174  & 22,966                  & 1,208                        \\
ReBEC (Brazil)  \cite{brazilclinicaltrials}    & 6,735    & 5,889                   & 846                         \\
CRIS (Korea) \cite{southkoreaclinicaltrials}                 & 8,953    & 8,428                   & 525                         \\
DRKS (German) \cite{germanclinicaltrials}                 & 15,693   & 13,789                  & 1,904                        \\
IRCT (Iran) \cite{iranianclinicaltrials}                & 37,782   & 34,097                  & 3,685                        \\
TCTR (Thailand)   \cite{thaiclinicaltrials}                  & 8,649    & 7,443                   & 1,206                        \\
LTR (Netherland) \cite{dutchclinicaltrials}             & 9,768    & 9,768                   & 0                           \\
PACTR (Africa) \cite{africanclinicaltrials}           & 4,047    & 3,848                   & 199                         \\
SLCTR (Sri Lanka)  \cite{srilankaclinicaltrials}            & 442     & 421                    & 21     \\
\midrule
Trial Papers (Embase \cite{embase} + PubMed \cite{pubmed}) & 1,113,207 & 1,113,207 & - \\
\bottomrule
\end{tabular}
}
\end{table}

\noindent 

\bibliographystyle{naturemag}
\bibliography{main}

\begin{thebibliography}{10}
\expandafter\ifx\csname url\endcsname\relax
  \def\url#1{\texttt{#1}}\fi
\expandafter\ifx\csname urlprefix\endcsname\relax\def\urlprefix{URL }\fi
\providecommand{\bibinfo}[2]{#2}
\providecommand{\eprint}[2][]{\url{#2}}

\bibitem{ling2023clinical}
\bibinfo{author}{Ling, A.~L.} \emph{et~al.}
\newblock \bibinfo{title}{Clinical trial links oncolytic immunoactivation to survival in glioblastoma}.
\newblock \emph{\bibinfo{journal}{Nature}} \textbf{\bibinfo{volume}{623}}, \bibinfo{pages}{157--166} (\bibinfo{year}{2023}).

\bibitem{heitmann2022covid}
\bibinfo{author}{Heitmann, J.~S.} \emph{et~al.}
\newblock \bibinfo{title}{A covid-19 peptide vaccine for the induction of sars-cov-2 t cell immunity}.
\newblock \emph{\bibinfo{journal}{Nature}} \textbf{\bibinfo{volume}{601}}, \bibinfo{pages}{617--622} (\bibinfo{year}{2022}).

\bibitem{hammond2024phase}
\bibinfo{author}{Hammond, T.~C.} \emph{et~al.}
\newblock \bibinfo{title}{A phase 1/2 clinical trial of invariant natural killer t cell therapy in moderate-severe acute respiratory distress syndrome}.
\newblock \emph{\bibinfo{journal}{Nature Communications}} \textbf{\bibinfo{volume}{15}}, \bibinfo{pages}{974} (\bibinfo{year}{2024}).

\bibitem{giamarellos2020activate}
\bibinfo{author}{Giamarellos-Bourboulis, E.~J.} \emph{et~al.}
\newblock \bibinfo{title}{Activate: randomized clinical trial of bcg vaccination against infection in the elderly}.
\newblock \emph{\bibinfo{journal}{Cell}} \textbf{\bibinfo{volume}{183}}, \bibinfo{pages}{315--323} (\bibinfo{year}{2020}).

\bibitem{gilbert2022immune}
\bibinfo{author}{Gilbert, P.~B.} \emph{et~al.}
\newblock \bibinfo{title}{Immune correlates analysis of the mrna-1273 covid-19 vaccine efficacy clinical trial}.
\newblock \emph{\bibinfo{journal}{Science}} \textbf{\bibinfo{volume}{375}}, \bibinfo{pages}{43--50} (\bibinfo{year}{2022}).

\bibitem{achiam2023gpt}
\bibinfo{author}{Achiam, J.} \emph{et~al.}
\newblock \bibinfo{title}{Gpt-4 technical report}.
\newblock \emph{\bibinfo{journal}{arXiv preprint arXiv:2303.08774}}  (\bibinfo{year}{2023}).

\bibitem{DBLP:conf/emnlp/0010X023}
\bibinfo{author}{Wang, Z.}, \bibinfo{author}{Xiao, C.} \& \bibinfo{author}{Sun, J.}
\newblock \bibinfo{title}{Autotrial: Prompting language models for clinical trial design}.
\newblock In \bibinfo{editor}{Bouamor, H.}, \bibinfo{editor}{Pino, J.} \& \bibinfo{editor}{Bali, K.} (eds.) \emph{\bibinfo{booktitle}{Proceedings of the 2023 Conference on Empirical Methods in Natural Language Processing, {EMNLP} 2023, Singapore, December 6-10, 2023}}, \bibinfo{pages}{12461--12472} (\bibinfo{publisher}{Association for Computational Linguistics}, \bibinfo{year}{2023}).

\bibitem{gao2020compose}
\bibinfo{author}{Gao, J.}, \bibinfo{author}{Xiao, C.}, \bibinfo{author}{Glass, L.~M.} \& \bibinfo{author}{Sun, J.}
\newblock \bibinfo{title}{Compose: Cross-modal pseudo-siamese network for patient trial matching}.
\newblock In \emph{\bibinfo{booktitle}{Proceedings of the 26th ACM SIGKDD international conference on knowledge discovery \& data mining}}, \bibinfo{pages}{803--812} (\bibinfo{year}{2020}).

\bibitem{wang2022trial2vec}
\bibinfo{author}{Wang, Z.} \& \bibinfo{author}{Sun, J.}
\newblock \bibinfo{title}{Trial2vec: Zero-shot clinical trial document similarity search using self-supervision}.
\newblock In \emph{\bibinfo{booktitle}{Findings of the Association for Computational Linguistics: EMNLP 2022}}, \bibinfo{pages}{6377--6390} (\bibinfo{year}{2022}).

\bibitem{gligorijevic2019optimizing}
\bibinfo{author}{Gligorijevic, J.} \emph{et~al.}
\newblock \bibinfo{title}{Optimizing clinical trials recruitment via deep learning}.
\newblock \emph{\bibinfo{journal}{Journal of the American Medical Informatics Association}} \textbf{\bibinfo{volume}{26}}, \bibinfo{pages}{1195--1202} (\bibinfo{year}{2019}).

\bibitem{zhang2020deepenroll}
\bibinfo{author}{Zhang, X.}, \bibinfo{author}{Xiao, C.}, \bibinfo{author}{Glass, L.~M.} \& \bibinfo{author}{Sun, J.}
\newblock \bibinfo{title}{Deepenroll: patient-trial matching with deep embedding and entailment prediction}.
\newblock In \emph{\bibinfo{booktitle}{Proceedings of the web conference 2020}}, \bibinfo{pages}{1029--1037} (\bibinfo{year}{2020}).

\bibitem{kim2021towards}
\bibinfo{author}{Kim, J.~H.} \emph{et~al.}
\newblock \bibinfo{title}{Towards clinical data-driven eligibility criteria optimization for interventional covid-19 clinical trials}.
\newblock \emph{\bibinfo{journal}{Journal of the American Medical Informatics Association}} \textbf{\bibinfo{volume}{28}}, \bibinfo{pages}{14--22} (\bibinfo{year}{2021}).

\bibitem{DBLP:journals/corr/abs-2307-14334}
\bibinfo{author}{Tu, T.} \emph{et~al.}
\newblock \bibinfo{title}{Towards generalist biomedical {AI}}.
\newblock \emph{\bibinfo{journal}{CoRR}} \textbf{\bibinfo{volume}{abs/2307.14334}} (\bibinfo{year}{2023}).

\bibitem{moor2023foundation}
\bibinfo{author}{Moor, M.} \emph{et~al.}
\newblock \bibinfo{title}{Foundation models for generalist medical artificial intelligence}.
\newblock \emph{\bibinfo{journal}{Nature}} \textbf{\bibinfo{volume}{616}}, \bibinfo{pages}{259--265} (\bibinfo{year}{2023}).

\bibitem{lu2024visual}
\bibinfo{author}{Lu, M.~Y.} \emph{et~al.}
\newblock \bibinfo{title}{A visual-language foundation model for computational pathology}.
\newblock \emph{\bibinfo{journal}{Nature Medicine}} \textbf{\bibinfo{volume}{30}}, \bibinfo{pages}{863--874} (\bibinfo{year}{2024}).

\bibitem{chen2024towards}
\bibinfo{author}{Chen, R.~J.} \emph{et~al.}
\newblock \bibinfo{title}{Towards a general-purpose foundation model for computational pathology}.
\newblock \emph{\bibinfo{journal}{Nature Medicine}} \textbf{\bibinfo{volume}{30}}, \bibinfo{pages}{850--862} (\bibinfo{year}{2024}).

\bibitem{cui2024scgpt}
\bibinfo{author}{Cui, H.} \emph{et~al.}
\newblock \bibinfo{title}{scgpt: toward building a foundation model for single-cell multi-omics using generative ai}.
\newblock \emph{\bibinfo{journal}{Nature Methods}} \bibinfo{pages}{1--11} (\bibinfo{year}{2024}).

\bibitem{huang2023visual}
\bibinfo{author}{Huang, Z.}, \bibinfo{author}{Bianchi, F.}, \bibinfo{author}{Yuksekgonul, M.}, \bibinfo{author}{Montine, T.~J.} \& \bibinfo{author}{Zou, J.}
\newblock \bibinfo{title}{A visual--language foundation model for pathology image analysis using medical twitter}.
\newblock \emph{\bibinfo{journal}{Nature medicine}} \textbf{\bibinfo{volume}{29}}, \bibinfo{pages}{2307--2316} (\bibinfo{year}{2023}).

\bibitem{xu2024whole}
\bibinfo{author}{Xu, H.} \emph{et~al.}
\newblock \bibinfo{title}{A whole-slide foundation model for digital pathology from real-world data}.
\newblock \emph{\bibinfo{journal}{Nature}} \bibinfo{pages}{1--8} (\bibinfo{year}{2024}).

\bibitem{DBLP:journals/corr/abs-2307-15051}
\bibinfo{author}{Jin, Q.} \emph{et~al.}
\newblock \bibinfo{title}{Matching patients to clinical trials with large language models}.
\newblock \emph{\bibinfo{journal}{ArXiv}}  (\bibinfo{year}{2023}).

\bibitem{yuan2023large}
\bibinfo{author}{Yuan, J.}, \bibinfo{author}{Tang, R.}, \bibinfo{author}{Jiang, X.} \& \bibinfo{author}{Hu, X.}
\newblock \bibinfo{title}{Large language models for healthcare data augmentation: An example on patient-trial matching}.
\newblock \emph{\bibinfo{journal}{arXiv preprint arXiv:2303.16756}}  (\bibinfo{year}{2023}).

\bibitem{DBLP:journals/corr/abs-2308-02180}
\bibinfo{author}{Wong, C.} \emph{et~al.}
\newblock \bibinfo{title}{Scaling clinical trial matching using large language models: {A} case study in oncology}.
\newblock \emph{\bibinfo{journal}{CoRR}} \textbf{\bibinfo{volume}{abs/2308.02180}} (\bibinfo{year}{2023}).

\bibitem{li2024llava}
\bibinfo{author}{Li, C.} \emph{et~al.}
\newblock \bibinfo{title}{Llava-med: Training a large language-and-vision assistant for biomedicine in one day}.
\newblock \emph{\bibinfo{journal}{Advances in Neural Information Processing Systems}} \textbf{\bibinfo{volume}{36}} (\bibinfo{year}{2024}).

\bibitem{chaves2024training}
\bibinfo{author}{Chaves, J. M.~Z.} \emph{et~al.}
\newblock \bibinfo{title}{Training small multimodal models to bridge biomedical competency gap: A case study in radiology imaging}.
\newblock \emph{\bibinfo{journal}{arXiv preprint arXiv:2403.08002}}  (\bibinfo{year}{2024}).

\bibitem{deyoung2021ms2}
\bibinfo{author}{DeYoung, J.}, \bibinfo{author}{Beltagy, I.}, \bibinfo{author}{van Zuylen, M.}, \bibinfo{author}{Kuehl, B.} \& \bibinfo{author}{Wang, L.~L.}
\newblock \bibinfo{title}{Ms2: Multi-document summarization of medical studies}.
\newblock \emph{\bibinfo{journal}{arXiv preprint arXiv:2104.06486}}  (\bibinfo{year}{2021}).

\bibitem{jiang2024trisum}
\bibinfo{author}{Jiang, P.} \emph{et~al.}
\newblock \bibinfo{title}{Trisum: Learning summarization ability from large language models with structured rationale}.
\newblock \emph{\bibinfo{journal}{arXiv preprint arXiv:2403.10351}}  (\bibinfo{year}{2024}).

\bibitem{jiang2023mistral}
\bibinfo{author}{Jiang, A.~Q.} \emph{et~al.}
\newblock \bibinfo{title}{Mistral 7b}.
\newblock \emph{\bibinfo{journal}{arXiv preprint arXiv:2310.06825}}  (\bibinfo{year}{2023}).

\bibitem{labrak2024biomistral}
\bibinfo{author}{Labrak, Y.} \emph{et~al.}
\newblock \bibinfo{title}{Biomistral: A collection of open-source pretrained large language models for medical domains}.
\newblock \emph{\bibinfo{journal}{arXiv preprint arXiv:2402.10373}}  (\bibinfo{year}{2024}).

\bibitem{anthropic2024claude}
\bibinfo{author}{Anthropic, A.}
\newblock \bibinfo{title}{The claude 3 model family: Opus, sonnet, haiku}.
\newblock \emph{\bibinfo{journal}{Claude-3 Model Card}}  (\bibinfo{year}{2024}).

\bibitem{roberts2021overview}
\bibinfo{author}{Roberts, K.}, \bibinfo{author}{Demner-Fushman, D.}, \bibinfo{author}{Voorhees, E.~M.}, \bibinfo{author}{Bedrick, S.} \& \bibinfo{author}{Hersh, W.~R.}
\newblock \bibinfo{title}{Overview of the trec 2021 clinical trials track}.
\newblock In \emph{\bibinfo{booktitle}{Proceedings of the thirtieth text retrieval conference (TREC 2021)}} (\bibinfo{year}{2021}).

\bibitem{koopman2016test}
\bibinfo{author}{Koopman, B.} \& \bibinfo{author}{Zuccon, G.}
\newblock \bibinfo{title}{A test collection for matching patients to clinical trials}.
\newblock In \emph{\bibinfo{booktitle}{Proceedings of the 39th International ACM SIGIR conference on Research and Development in Information Retrieval}}, \bibinfo{pages}{669--672} (\bibinfo{year}{2016}).

\bibitem{touvron2023llama}
\bibinfo{author}{Touvron, H.} \emph{et~al.}
\newblock \bibinfo{title}{Llama 2: Open foundation and fine-tuned chat models}.
\newblock \emph{\bibinfo{journal}{arXiv preprint arXiv:2307.09288}}  (\bibinfo{year}{2023}).

\bibitem{DBLP:journals/bib/LuoSXQZPL22}
\bibinfo{author}{Luo, R.} \emph{et~al.}
\newblock \bibinfo{title}{Biogpt: generative pre-trained transformer for biomedical text generation and mining}.
\newblock \emph{\bibinfo{journal}{Briefings Bioinform.}} \textbf{\bibinfo{volume}{23}} (\bibinfo{year}{2022}).

\bibitem{singhal2023large}
\bibinfo{author}{Singhal, K.} \emph{et~al.}
\newblock \bibinfo{title}{Large language models encode clinical knowledge}.
\newblock \emph{\bibinfo{journal}{Nature}} \textbf{\bibinfo{volume}{620}}, \bibinfo{pages}{172--180} (\bibinfo{year}{2023}).

\bibitem{chen2023meditron}
\bibinfo{author}{Chen, Z.} \emph{et~al.}
\newblock \bibinfo{title}{Meditron-70b: Scaling medical pretraining for large language models}.
\newblock \emph{\bibinfo{journal}{arXiv preprint arXiv:2311.16079}}  (\bibinfo{year}{2023}).

\bibitem{van2024adapted}
\bibinfo{author}{Van~Veen, D.} \emph{et~al.}
\newblock \bibinfo{title}{Adapted large language models can outperform medical experts in clinical text summarization}.
\newblock \emph{\bibinfo{journal}{Nature Medicine}} \bibinfo{pages}{1--9} (\bibinfo{year}{2024}).

\bibitem{tayebi2024large}
\bibinfo{author}{Tayebi~Arasteh, S.} \emph{et~al.}
\newblock \bibinfo{title}{Large language models streamline automated machine learning for clinical studies}.
\newblock \emph{\bibinfo{journal}{Nature Communications}} \textbf{\bibinfo{volume}{15}}, \bibinfo{pages}{1603} (\bibinfo{year}{2024}).

\bibitem{DBLP:journals/corr/abs-2311-16452}
\bibinfo{author}{Nori, H.} \emph{et~al.}
\newblock \bibinfo{title}{Can generalist foundation models outcompete special-purpose tuning? case study in medicine}.
\newblock \emph{\bibinfo{journal}{CoRR}} \textbf{\bibinfo{volume}{abs/2311.16452}} (\bibinfo{year}{2023}).

\bibitem{ouyang2022training}
\bibinfo{author}{Ouyang, L.} \emph{et~al.}
\newblock \bibinfo{title}{Training language models to follow instructions with human feedback}.
\newblock \emph{\bibinfo{journal}{Advances in Neural Information Processing Systems}} \textbf{\bibinfo{volume}{35}}, \bibinfo{pages}{27730--27744} (\bibinfo{year}{2022}).

\bibitem{lin2023generating}
\bibinfo{author}{Lin, Z.}, \bibinfo{author}{Trivedi, S.} \& \bibinfo{author}{Sun, J.}
\newblock \bibinfo{title}{Generating with confidence: Uncertainty quantification for black-box large language models}.
\newblock \emph{\bibinfo{journal}{arXiv preprint arXiv:2305.19187}}  (\bibinfo{year}{2023}).

\bibitem{semnani2023wikichat}
\bibinfo{author}{Semnani, S.}, \bibinfo{author}{Yao, V.}, \bibinfo{author}{Zhang, H.} \& \bibinfo{author}{Lam, M.}
\newblock \bibinfo{title}{{WikiChat}: Stopping the hallucination of large language model chatbots by few-shot grounding on wikipedia}.
\newblock In \emph{\bibinfo{booktitle}{Findings of the Association for Computational Linguistics: EMNLP 2023}}, \bibinfo{pages}{2387--2413} (\bibinfo{year}{2023}).

\bibitem{hu2021lora}
\bibinfo{author}{Hu, E.~J.} \emph{et~al.}
\newblock \bibinfo{title}{{LoRA}: Low-rank adaptation of large language models}.
\newblock In \emph{\bibinfo{booktitle}{International Conference on Learning Representations}} (\bibinfo{year}{2021}).

\bibitem{lewis2020retrieval}
\bibinfo{author}{Lewis, P.} \emph{et~al.}
\newblock \bibinfo{title}{Retrieval-augmented generation for knowledge-intensive nlp tasks}.
\newblock \emph{\bibinfo{journal}{Advances in Neural Information Processing Systems}} \textbf{\bibinfo{volume}{33}}, \bibinfo{pages}{9459--9474} (\bibinfo{year}{2020}).

\bibitem{cochranecentral}
\bibinfo{author}{Collaboration, C.} \emph{et~al.}
\newblock \bibinfo{title}{Cochrane central register of controlled trials (central)} (\bibinfo{year}{2014}).

\bibitem{clinicaltrialsgov}
\bibinfo{author}{Bergeris, A.}, \bibinfo{author}{Ide, N.~C.} \& \bibinfo{author}{Tse, T.}
\newblock \bibinfo{title}{Clinicaltrials. gov}  (\bibinfo{year}{2005}).

\bibitem{wallace2021generating}
\bibinfo{author}{Wallace, B.~C.}, \bibinfo{author}{Saha, S.}, \bibinfo{author}{Soboczenski, F.} \& \bibinfo{author}{Marshall, I.~J.}
\newblock \bibinfo{title}{Generating (factual?) narrative summaries of rcts: Experiments with neural multi-document summarization}.
\newblock \emph{\bibinfo{journal}{AMIA Summits on Translational Science Proceedings}} \textbf{\bibinfo{volume}{2021}}, \bibinfo{pages}{605} (\bibinfo{year}{2021}).

\bibitem{DBLP:conf/iclr/LoshchilovH19}
\bibinfo{author}{Loshchilov, I.} \& \bibinfo{author}{Hutter, F.}
\newblock \bibinfo{title}{Decoupled weight decay regularization}.
\newblock In \emph{\bibinfo{booktitle}{7th International Conference on Learning Representations, {ICLR} 2019, New Orleans, LA, USA, May 6-9, 2019}} (\bibinfo{publisher}{OpenReview.net}, \bibinfo{year}{2019}).
\newblock \urlprefix\url{https://openreview.net/forum?id=Bkg6RiCqY7}.

\bibitem{rajbhandari2020zero}
\bibinfo{author}{Rajbhandari, S.}, \bibinfo{author}{Rasley, J.}, \bibinfo{author}{Ruwase, O.} \& \bibinfo{author}{He, Y.}
\newblock \bibinfo{title}{Zero: Memory optimizations toward training trillion parameter models}.
\newblock In \emph{\bibinfo{booktitle}{SC20: International Conference for High Performance Computing, Networking, Storage and Analysis}}, \bibinfo{pages}{1--16} (\bibinfo{organization}{IEEE}, \bibinfo{year}{2020}).

\bibitem{dao2023flashattention}
\bibinfo{author}{Dao, T.}
\newblock \bibinfo{title}{Flashattention-2: Faster attention with better parallelism and work partitioning}.
\newblock \emph{\bibinfo{journal}{arXiv preprint arXiv:2307.08691}}  (\bibinfo{year}{2023}).

\bibitem{jsonformer}
\bibinfo{author}{1rgs}.
\newblock \bibinfo{title}{Jsonformer: A bulletproof way to generate structured json from language models} (\bibinfo{year}{2023}).

\bibitem{chictr}
\bibinfo{author}{Wu, T.} \emph{et~al.}
\newblock \bibinfo{title}{Chinese clinical trial registry: mission, responsibility and operation}.
\newblock \emph{\bibinfo{journal}{Journal of evidence-based medicine}} \textbf{\bibinfo{volume}{4}}, \bibinfo{pages}{165--167} (\bibinfo{year}{2011}).

\bibitem{euclinicaltrialsregister}
\bibinfo{author}{Egger, G.~F.} \emph{et~al.}
\newblock \bibinfo{title}{European union clinical trials register: on the way to more transparency of clinical trial data}.
\newblock \emph{\bibinfo{journal}{Expert Review of Clinical Pharmacology}} \textbf{\bibinfo{volume}{6}}, \bibinfo{pages}{457--459} (\bibinfo{year}{2013}).

\bibitem{japanrctportal}
\bibinfo{author}{Shiokawa, T.}
\newblock \bibinfo{title}{Background, introduction and activity of the japan primary registries network}.
\newblock \emph{\bibinfo{journal}{Journal of Evidence-Based Medicine}} \textbf{\bibinfo{volume}{2}}, \bibinfo{pages}{41--43} (\bibinfo{year}{2009}).

\bibitem{anzctr}
\bibinfo{author}{Askie, L.~M.}
\newblock \bibinfo{title}{Australian new zealand clinical trials registry: history and growth}.
\newblock \emph{\bibinfo{journal}{Journal of Evidence-Based Medicine}} \textbf{\bibinfo{volume}{4}}, \bibinfo{pages}{185--187} (\bibinfo{year}{2011}).

\bibitem{isrctnregistry}
\bibinfo{author}{Faure, H.} \& \bibinfo{author}{Hrynaszkiewicz, I.}
\newblock \bibinfo{title}{The isrctn register: achievements and challenges 8 years on}.
\newblock \emph{\bibinfo{journal}{Journal of evidence-based medicine}} \textbf{\bibinfo{volume}{4}}, \bibinfo{pages}{188--192} (\bibinfo{year}{2011}).

\bibitem{brazilclinicaltrials}
\bibinfo{author}{Laguardia, J.} \emph{et~al.}
\newblock \bibinfo{title}{Brazilian clinical trials registry and the challenges for clinical research governance}.
\newblock \emph{\bibinfo{journal}{Journal of Evidence-Based Medicine}} \textbf{\bibinfo{volume}{4}}, \bibinfo{pages}{156--160} (\bibinfo{year}{2011}).

\bibitem{southkoreaclinicaltrials}
\bibinfo{author}{Park, H.-Y.}
\newblock \bibinfo{title}{Primary registry of the who international clinical trial registry platform: Clinical research information service (cris)}.
\newblock \emph{\bibinfo{journal}{Journal of the Korean Medical Association}} \textbf{\bibinfo{volume}{54}}, \bibinfo{pages}{92--97} (\bibinfo{year}{2011}).

\bibitem{germanclinicaltrials}
\bibinfo{author}{Hasselblatt, H.}, \bibinfo{author}{Dreier, G.}, \bibinfo{author}{Antes, G.} \& \bibinfo{author}{Schumacher, M.}
\newblock \bibinfo{title}{The german clinical trials register: challenges and chances of implementing a bilingual registry}.
\newblock \emph{\bibinfo{journal}{Journal of Evidence-Based Medicine}} \textbf{\bibinfo{volume}{2}}, \bibinfo{pages}{36--40} (\bibinfo{year}{2009}).

\bibitem{iranianclinicaltrials}
\bibinfo{author}{Solaymani-Dodaran, M.}, \bibinfo{author}{Ostovar, A.}, \bibinfo{author}{Khalili, D.} \& \bibinfo{author}{Vasei, M.}
\newblock \bibinfo{title}{Iranian registry of clinical trials: path and challenges from conception to a world health organization primary register}.
\newblock \emph{\bibinfo{journal}{Journal of Evidence-Based Medicine}} \textbf{\bibinfo{volume}{2}}, \bibinfo{pages}{32--35} (\bibinfo{year}{2009}).

\bibitem{thaiclinicaltrials}
\bibinfo{author}{Tulvatana, W.}, \bibinfo{author}{Kulvichit, K.}, \bibinfo{author}{Thinkhamrop, B.} \& \bibinfo{author}{Tatsanavivat, P.}
\newblock \bibinfo{title}{Thai clinical trials registry}.
\newblock \emph{\bibinfo{journal}{Journal of Evidence-Based Medicine}} \textbf{\bibinfo{volume}{4}}, \bibinfo{pages}{182--184} (\bibinfo{year}{2011}).

\bibitem{dutchclinicaltrials}
\bibinfo{author}{Driessen, M.} \emph{et~al.}
\newblock \bibinfo{title}{The dutch nationwide trauma registry: the value of capturing all acute trauma admissions}.
\newblock \emph{\bibinfo{journal}{Injury}} \textbf{\bibinfo{volume}{51}}, \bibinfo{pages}{2553--2559} (\bibinfo{year}{2020}).

\bibitem{africanclinicaltrials}
\bibinfo{author}{Abrams, A.} \& \bibinfo{author}{Siegfried, N.}
\newblock \bibinfo{title}{The pan african clinical trials registry: year one data analysis of the only african member of the world health organization network of primary registries}.
\newblock \emph{\bibinfo{journal}{Journal of Evidence-Based Medicine}} \textbf{\bibinfo{volume}{3}}, \bibinfo{pages}{195--200} (\bibinfo{year}{2010}).

\bibitem{srilankaclinicaltrials}
\bibinfo{author}{Ranawaka, U.~K.} \& \bibinfo{author}{Goonaratna, C.}
\newblock \bibinfo{title}{The sri lanka clinical trials registry--moving forward}.
\newblock \emph{\bibinfo{journal}{Journal of Evidence-Based Medicine}} \textbf{\bibinfo{volume}{4}}, \bibinfo{pages}{179--181} (\bibinfo{year}{2011}).

\bibitem{embase}
\bibinfo{author}{{Elsevier Science}}.
\newblock \bibinfo{title}{Embase [electronic database]}.
\newblock \bibinfo{howpublished}{Electronic Database} (\bibinfo{year}{1974}).
\newblock \bibinfo{note}{Produced by Elsevier Science, Amsterdam, The Netherlands}.

\bibitem{pubmed}
\bibinfo{author}{Canese, K.} \& \bibinfo{author}{Weis, S.}
\newblock \bibinfo{title}{Pubmed: the bibliographic database}.
\newblock \emph{\bibinfo{journal}{The NCBI handbook}} \textbf{\bibinfo{volume}{2}} (\bibinfo{year}{2013}).

\end{thebibliography}

\end{document}